\documentclass[11pt]{article}

\usepackage[final]{acl}

\usepackage{times}
\usepackage{latexsym}

\usepackage[T1]{fontenc}

\usepackage[utf8]{inputenc}

\usepackage{microtype}

\usepackage{inconsolata}

\usepackage{graphicx}
\usepackage[table]{xcolor}
\usepackage{amsmath}
\usepackage{amssymb}
\usepackage{bbm}
\usepackage{booktabs}
\usepackage{multirow}
\usepackage{array}
\usepackage{makecell}
\usepackage{pifont}
\usepackage{stfloats}
\usepackage{cuted}
\usepackage{capt-of}
\usepackage{placeins}
\newcommand{\cmark}{\textcolor{green!60!black}{\ding{51}}}
\newcommand{\xmark}{\textcolor{red!70!black}{\ding{55}}}

\title{MEMOBench: A Process Level Memory Benchmark for Robotic Manipulation}

\author{
  \textbf{Haiyang Sun\textsuperscript{*1}},
  \textbf{Haoxiao Wang\textsuperscript{*1}},
  \textbf{Junming Chen\textsuperscript{*1}},
  \textbf{Weicheng Fang\textsuperscript{1}},
  \textbf{Zihao Su\textsuperscript{1}},
\\
  \textbf{Jingkun Yi\textsuperscript{1}},
  \textbf{Wenyou Yi\textsuperscript{1}},
  \textbf{Hao Chen\textsuperscript{1}},
  \textbf{Zhou Zhao\textsuperscript{1,}\textsuperscript{\dag}}
\\
\\
  \textsuperscript{1}Zhejiang University, Hangzhou, China
\\
  \small{
    \texttt{\{sunsealucky, 22551003, 3240106190, 3240104144\}@zju.edu.cn}
  }
\\
  \small{
    \texttt{\{3240102078, 22471163, 22651324\}@zju.edu.cn,}
    \texttt{c3236455482@gmail.com}
  }
\\
  \small{
    \texttt{zhaozhou@zju.edu.cn}
  }
\\
  \small{
    \textsuperscript{*}Equal contribution
    \textsuperscript{\dag}Corresponding author
  }
}

\begin{document}
\maketitle

\begin{abstract}

  Robotic manipulation often requires acting on information that is no longer visible, yet Vision-Language-Action policies are usually evaluated when the current observation largely determines the next action. Existing robotic memory benchmarks expose this gap, but they still rely mainly on final task success and therefore conflate forgetting with manipulation failure. We present \textbf{MEMOBench}, a benchmark for process level memory evaluation in robotic manipulation. MEMOBench includes 30 history dependent tasks, 1{,}500 expert demonstrations, and 4{,}200 executable checkpoint instances from 84 templates. Each checkpoint pairs coarse to fine language with a simulator predicate and labels one memory operation: Storage, Update, or Compression. These annotations define Memory Storage Rate, Memory Update Rate, and Memory Compression Rate, which measure memory fidelity alongside task success. Across standard and memory augmented VLA policies, the strongest memory module baseline reaches only 31.9\% average success rate, and high storage often coexists with weak update and compression. Checkpoint language also supervises semantic, contrastive, and framewise memory alignment objectives, yielding modest gains across different memory operations. MEMOBench provides a diagnostic evaluation suite and training supervision for memory grounded robotic policies. The project page is available at \url{https://github.com/Collab-Gen/MEMOBench}.

\end{abstract}

\section{Introduction}

\begin{figure*}[!t]
  \centering
  \begin{minipage}[c]{0.76\textwidth}
    \centering
    \includegraphics[width=\linewidth]{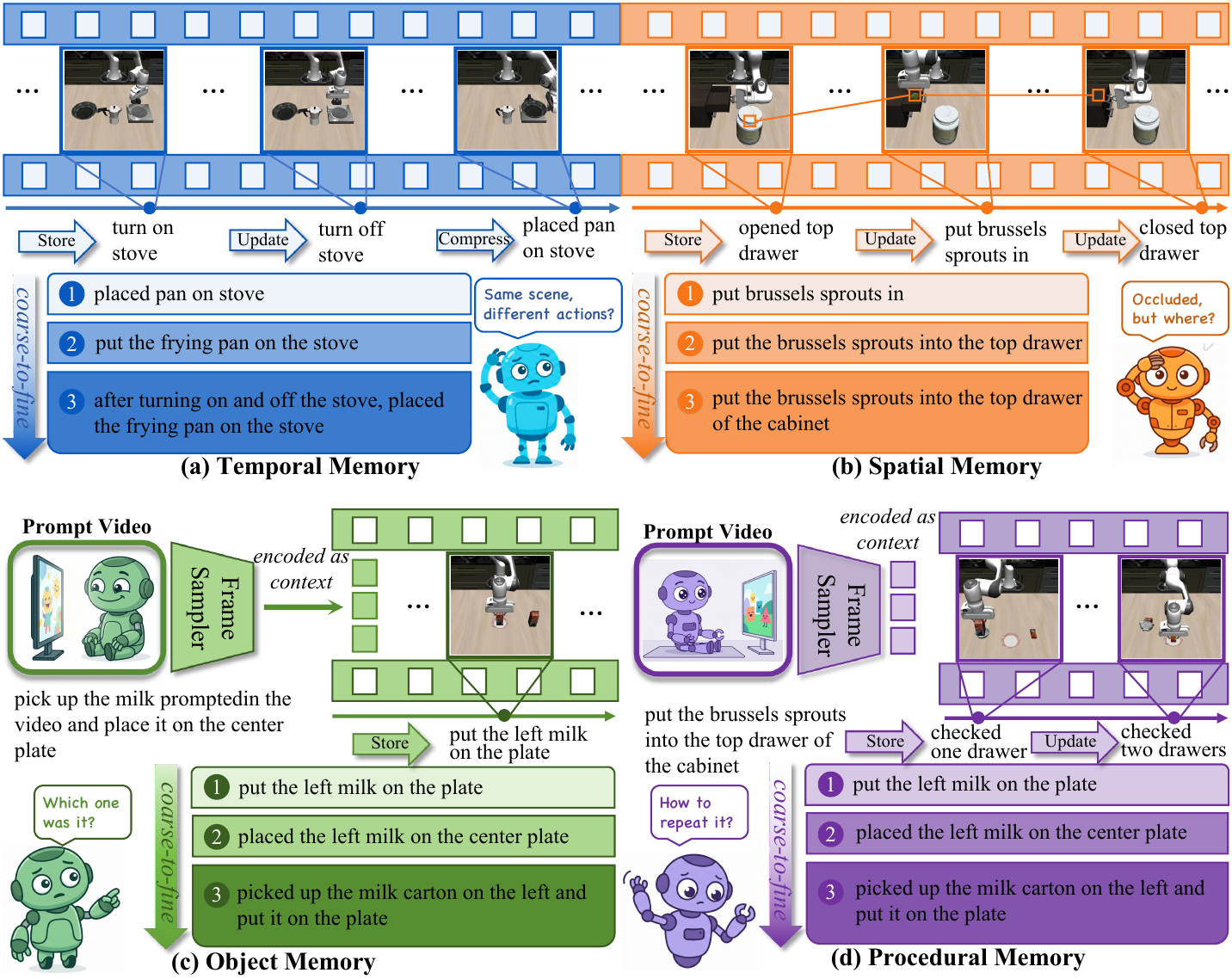}
  \end{minipage}%
  \begin{minipage}[c]{0.24\textwidth}
    \centering
    \includegraphics[width=\linewidth]{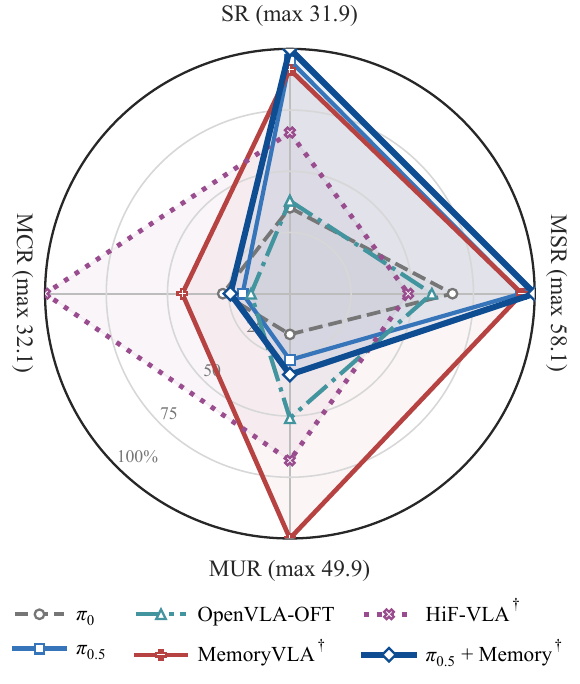}\\[-0.3em]
    \includegraphics[width=\linewidth]{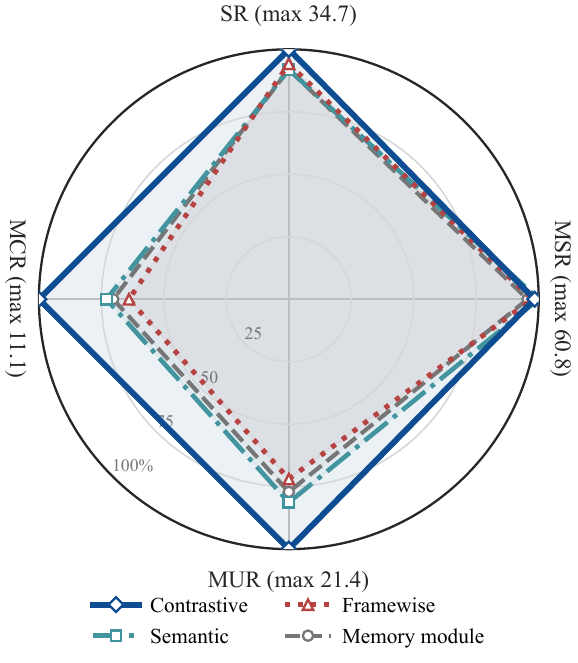}
  \end{minipage}
  \caption{Overview of MEMOBench, including history dependent manipulation tasks, process level checkpoint annotations, and radar summaries for model evaluation and memory alignment.}
  \label{fig:teaser}
\end{figure*}

Robotic manipulation often requires a policy to act on information that is absent from the current observation. A household robot may need to place an object at a location named earlier, turn off a stove after a delay, or repeat an action a specified number of times. These settings break the Markov assumption because the correct action depends on stored references, changed states, or accumulated events. Recent Vision-Language-Action (VLA) models, including $\pi_0$~\cite{black2026pi0visionlanguageactionflowmodel}, $\pi_{0.5}$~\cite{intelligence2025pi05visionlanguageactionmodelopenworld}, OpenVLA~\cite{kim2024openvlaopensourcevisionlanguageactionmodel}, and Octo~\cite{octomodelteam2024octoopensourcegeneralistrobot}, have advanced robotic manipulation, but prior analysis reports only 0 to 5\% success when such policies face memory dependent tasks~\cite{chung2025rethinkingprogressionmemorystate}. This gap calls for benchmarks that evaluate memory as a core capability. Figure~\ref{fig:teaser} previews the MEMOBench task setting, checkpoint diagnosis, and result summaries that address this gap.

Existing robotic memory benchmarks expose the problem, but they still make memory failures difficult to diagnose~\cite{fang2025sam2actintegratingvisualfoundation, cherepanov2026memorybenchmarkrobots, dai2026robommebenchmarkingunderstandingmemory, chen2026rmbenchmemorydependentroboticmanipulation, han2025robocerebralargescalebenchmarklonghorizon}. Many tasks use structurally simple targets, such as blocks or buttons, and most demonstrations do not mark when information should be stored, revised, or consolidated. Evaluation also relies mainly on final task success, which mixes memory failure with manipulation failure. A failed rollout therefore does not reveal whether the policy failed to store relevant information, update stale state, compress history, or execute the right action after remembering correctly. A benchmark for memory grounded manipulation should make these operations observable and measure them separately from task completion.

We propose \textbf{MEMOBench}, a benchmark for process level memory evaluation in robotic manipulation. MEMOBench contains 30 history dependent tasks and 1{,}500 expert demonstrations over household objects and fixtures. Its 84 executable checkpoint templates yield 4{,}200 demonstration level checkpoint instances, each with coarse to fine language, a simulator predicate, and one operation label: Storage for newly observed information, Update for changed state, or Compression for accumulated history. These checkpoints support Memory Storage Rate, Memory Update Rate, and Memory Compression Rate, which measure memory fidelity alongside task success without reading model internals.

MEMOBench shows that current VLA policies remain far from reliable memory grounded manipulation. We evaluate six models on MEMOBench. The strongest memory module baseline reaches 31.9\% average success rate under its best replan setting, and high storage rates often coexist with weak update and compression. We also use checkpoint language as supervision for semantic, contrastive, and framewise memory alignment on the same $\pi_{0.5}$ memory module backbone~\cite{dai2026robommebenchmarkingunderstandingmemory}. Alignment gives modest gains and benefits different operations depending on the objective, showing that MEMOBench can serve as both an evaluation suite and a source of supervision.

Our contributions are summarized as follows:
\begin{itemize}
    \item We introduce MEMOBench, a robotic manipulation memory benchmark with process level memory operation annotations, covering 30 tasks, 1{,}500 demonstrations, and 4{,}200 checkpoint instances.
    \item We define executable checkpoints and Memory Storage, Update, and Compression Rates to evaluate memory fidelity beyond final task success.
    \item We evaluate standard and memory augmented VLA policies, showing that current policies often store task information but struggle to update and compress it over time.
    \item We use checkpoint language for controlled memory alignment, showing that annotation based supervision can benefit different memory operations.
\end{itemize}

\section{Related Work}

\textbf{General Robotic Manipulation Benchmarks.}
Robot intelligence research has long emphasized manipulation intelligence, the skill of reliably manipulating objects according to tasks~\cite{liu2018robotintelligencerealworld}.
General manipulation benchmarks have driven policies that map current observations and language instructions to actions.
RLBench~\cite{james2019rlbenchrobotlearningbenchmark}, CALVIN~\cite{mees2022calvinbenchmarklanguageconditionedpolicy}, LIBERO~\cite{liu2023liberobenchmarkingknowledgetransfer}, ManiSkill3~\cite{tao2025maniskill3gpuparallelizedrobotics}, and BridgeData~V2~\cite{walke2024bridgedatav2datasetrobot} cover diverse skills, task chains, simulation platforms, and real trajectories.
However, they usually assume that the current observation and instruction contain the information needed for the next action, and they score final completion rather than whether a reference, state change, or event count was stored before later use.
MEMOBench targets this missing setting by making every task history dependent and by evaluating memory alongside manipulation success.

\textbf{Benchmarks for Tasks Requiring History.}
Recent history dependent and long horizon benchmarks move beyond single observation evaluation by requiring policies to reason over prior observations and actions.
MemoryBench~\cite{fang2025sam2actintegratingvisualfoundation} introduced three tasks that require history, showing an early path toward explicit memory evaluation in manipulation.
MIKASA-Robo~\cite{cherepanov2026memorybenchmarkrobots} scales memory evaluation to 32 ManiSkill3 tasks across object, spatial, sequential, and capacity categories.
LIBERO-Mem~\cite{chung2025rethinkingprogressionmemorystate} exposes the weakness of VLA policies on object centered memory tasks, reporting 0 to 5\% success across 10 tasks.
RoboMME~\cite{dai2026robommebenchmarkingunderstandingmemory} offers a broad prior taxonomy by grounding 16 tasks in the Atkinson-Shiffrin memory model~\cite{ATKINSON196889}, collecting 1{,}600 demonstrations, and evaluating 14 model variants.
RMBench~\cite{chen2026rmbenchmemorydependentroboticmanipulation} formalizes task memory demand with Task Memory Complexity across 9 dual arm tasks.
RoboCerebra~\cite{han2025robocerebralargescalebenchmarklonghorizon} studies the temporal dimension at larger scale through 100 long horizon tasks with Memory Exploration and Memory Execution subtasks.
RoboMemArena~\cite{lei2026robomemarenacomprehensivechallengingrobotic} further raises the difficulty with 26 simulation tasks, 5 real robot tasks, and keyframe and subtask supervision.
Together, these benchmarks establish memory as a meaningful bottleneck for robot policies.
Machine memory research makes the same point from the cognitive side, arguing that intelligent systems need explicit storage structures that support dynamic updates and associative retrieval~\cite{zheng2025machinememoryintelligence}.
However, they still make memory failures difficult to localize: annotations usually identify subtasks or keyframes rather than explicit memory operations, and evaluation is dominated by final or stage success rates that mix memory fidelity with motor execution.
MEMOBench addresses this gap with household object scenarios, coarse to fine checkpoint language, executable checkpoint predicates, and process level metrics for Storage, Update, and Compression.

\section{MEMOBench Benchmark}
\label{sec:benchmark}
\subsection{Memory Taxonomy}

MEMOBench uses a two layer taxonomy that separates \emph{what} a task requires from \emph{how} the remembered state changes during execution. The task type layer follows prior cognitive memory benchmarks~\cite{dai2026robommebenchmarkingunderstandingmemory, cherepanov2026memorybenchmarkrobots} and assigns each task to the dominant content that must remain available beyond the current observation:
\begin{itemize}
    \item \textbf{Temporal Memory} (\emph{when}): retaining event counts, temporal order, and transition conditions across timesteps.
    \item \textbf{Spatial Memory} (\emph{where}): retaining object locations and spatial relationships when the current visual observation is insufficient.
    \item \textbf{Object Memory} (\emph{what}): preserving object identity and referential consistency across visually similar instances or earlier language cues.
    \item \textbf{Procedural Memory} (\emph{how}): retaining demonstrated motion patterns or manipulation sequences as constraints on later actions.
\end{itemize}

The memory operation layer assigns each executable checkpoint to the change in task relevant memory state. Here, $\mathcal{M}$ denotes the information implied by the checkpoint annotation, not a direct readout from model hidden states. We use three operation labels:
\begin{itemize}
    \item \textbf{Storage} ($\varnothing \to \mathcal{M}$): encoding newly observed task information.
    \item \textbf{Update} ($\mathcal{M} \to \mathcal{M}'$): revising an existing memory entry after the external state changes, following the intuition of memory reconsolidation~\cite{nader_fear_2000}.
    \item \textbf{Compression} ($\mathcal{M} \to \hat{\mathcal{M}}$): aggregating accumulated information into a compact task representation, following the intuition of chunking in working memory~\cite{miller1956magical}.
\end{itemize}
The operation labels are mutually exclusive at each checkpoint. Update covers state changes, including deletion, while Compression covers accumulated information. Retrieval is not annotated because it may occur at any decision step. A blind relabeling study with three independent annotators confirms that these labels can be applied consistently, with Fleiss' $\kappa=0.799$ (Appendix~\ref{sec:appendix_annotator_agreement}). Together, the two taxonomy layers let MEMOBench report both \emph{which} memory type is difficult and \emph{which} operation fails. Table~\ref{tab:dataset_comparison} situates this design relative to existing manipulation benchmarks.

\begin{table*}[t]
\centering
\small
\renewcommand{\arraystretch}{1.12}
\setlength{\tabcolsep}{4.5pt}
\resizebox{\textwidth}{!}{
\begin{tabular}{@{}l c ccc cc ccc rrr@{}}
\toprule
 & & \multicolumn{3}{c}{\textsc{Task Properties}} & \multicolumn{2}{c}{\textsc{Inputs}} & \multicolumn{3}{c}{\textsc{Annotations}} & \multicolumn{3}{c}{\textsc{Scale}} \\
\cmidrule(lr){3-5} \cmidrule(lr){6-7} \cmidrule(lr){8-10} \cmidrule(lr){11-13}
\textbf{Benchmark}
 & \makecell[c]{\textbf{Mem.}\\\textbf{Type}}
 & \makecell[c]{\textbf{Non-}\\\textbf{Mkv.}}
 & \makecell[c]{\textbf{Part.}\\\textbf{Obs.}}
 & \makecell[c]{\textbf{Dyn.}\\\textbf{Scene}}
 & \textbf{Vid.}
 & \textbf{Lang.}
 & \textbf{Subgoal}
 & \textbf{Keyframe}
 & \makecell[c]{\textbf{Mem.}\\\textbf{Op.}}
 & \textbf{\#Task}
 & \textbf{\#Demo}
 & \textbf{Avg. Len.} \\
\midrule
\multicolumn{13}{@{}l}{\textit{General purpose manipulation benchmarks}} \\[1pt]
RLBench~\cite{james2019rlbenchrobotlearningbenchmark}       & T       & \xmark & \xmark & \xmark & \xmark & \cmark & \xmark & \cmark & \xmark & 18  & 1{,}800  & 137 \\
CALVIN~\cite{mees2022calvinbenchmarklanguageconditionedpolicy}        & T       & \xmark & \xmark & \xmark & \xmark & \cmark & \cmark & \cmark & \xmark & 34  & 1{,}000  & 584 \\
LIBERO~\cite{liu2023liberobenchmarkingknowledgetransfer}        & T       & \xmark & \xmark & \xmark & \xmark & \cmark & \xmark & \xmark & \xmark & 130 & 6{,}500  & 162 \\
RoboCerebra~\cite{han2025robocerebralargescalebenchmarklonghorizon}   & T       & \xmark & \xmark & \cmark & \xmark & \cmark & \cmark & \xmark & \xmark & 100 & 1{,}000  & 2{,}972 \\
\addlinespace[3pt]
\multicolumn{13}{@{}l}{\textit{Memory focused manipulation benchmarks}} \\[1pt]
MemoryBench~\cite{fang2025sam2actintegratingvisualfoundation}   & T+S     & \cmark & \xmark & \xmark & \xmark & \cmark & \xmark & \cmark & \xmark & 3   & 300      & 312 \\
MIKASA-Robo~\cite{cherepanov2026memorybenchmarkrobots}   & S+O+Seq+Cap   & \cmark & \cmark & \cmark & \xmark & \xmark & \xmark & \xmark & \xmark & 32  & 1{,}250  & 72 \\
RoboMME~\cite{dai2026robommebenchmarkingunderstandingmemory}       & T+S+O+P & \cmark & \cmark & \cmark & \cmark & \cmark & \cmark & \cmark & \xmark & 16  & 1{,}600  & 481 \\
RoboMemArena~\cite{lei2026robomemarenacomprehensivechallengingrobotic}       & T+S+O+Seq & \cmark & \cmark & \cmark & \cmark & \cmark & \cmark & \cmark & \xmark & 26  & 2{,}600  & 1{,}076 \\
\midrule
\rowcolor{gray!8}
\textbf{MEMOBench (Ours)} & \textbf{T+S+O+P} & \cmark & \cmark & \cmark & \cmark & \cmark & \cmark & \cmark & \cmark & \textbf{30} & \textbf{1{,}500} & \textbf{618} \\
\bottomrule
\end{tabular}}
\caption{
Comparison with existing manipulation benchmarks. Memory types: T = temporal, S = spatial, O = object, P = procedural, Seq = sequential, Cap = memory capacity. RLBench reports the commonly evaluated 18 task subset. RoboMemArena reports its simulation benchmark only, excluding 5 real robot tasks. The final column reports average trajectory length when available. MEMOBench is the first robotic manipulation memory benchmark with process level memory operation annotations, diagnosing \emph{which} memory operation fails rather than only \emph{whether} the task succeeds.
}
\label{tab:dataset_comparison}
\end{table*}

\subsection{Task Suite}

MEMOBench contains 30 history dependent manipulation tasks implemented as BDDL task specifications on the LIBERO/robosuite stack~\cite{liu2023liberobenchmarkingknowledgetransfer}. Each task has 50 expert demonstrations, giving 1{,}500 demonstrations over 29 object classes and five fixture classes. The tasks use household settings in which object identity, spatial location, interaction history, or demonstrated routines must be preserved after they are no longer visible.

Table~\ref{tab:task_suite} in Appendix~\ref{sec:appendix_task_suite} summarizes the task families and operation coverage. Temporal tasks preserve event order, state transitions, or repeated interactions. Spatial tasks maintain object locations across drawer events. Object tasks isolate referential memory among similar instances. Procedural tasks require reproducing an observed routine. The suite mixes operation profiles so each family can be interpreted through both memory type and executable checkpoints.

\subsection{Memory Checkpoint Annotations}

Each task $\tau$ specifies memory checkpoints $\mathcal{C}_\tau = \{c_1,\ldots,c_K\}$. A checkpoint contains a coarse to fine language annotation, a memory operation label, and an executable subgoal predicate. The three language levels range from concise phrases such as ``put tomato in'' to grounded descriptions such as ``put the tomato into the top drawer of the cabinet''.

Across the 30 tasks, MEMOBench defines 84 checkpoint templates. With 50 demonstrations per task, these templates yield 4{,}200 checkpoint instances at the demonstration level. At the default annotation granularity used in evaluation, the suite contains 1{,}750 Storage instances, 1{,}600 Update instances, and 850 Compression instances. The distribution is intentionally not uniform: object reference tasks primarily stress Storage, spatial drawer tasks exercise all three operations, and procedural tasks emphasize the transition from observed routine to executable action sequence.

The predicate field makes each checkpoint directly verifiable in the simulator. Rather than reading internal memory states, MEMOBench evaluates whether the environment satisfies the state implied by the checkpoint. Checkpoints are ordered by the task script, and order sensitive analyses report prefix completion, where checkpoint $c_k$ counts as progress only when $(c_1,\ldots,c_k)$ has been completed. This sequence aware view prevents an isolated later predicate from being interpreted as valid task progress. These predicates connect the annotation layer to the process level metrics in Section~\ref{sec:metrics}. Appendix~\ref{sec:appendix_data_collection} details data collection and quality control.

\subsection{Memory Alignment from Annotations}

\begin{figure*}[t]
  \centering
  \includegraphics[width=\textwidth]{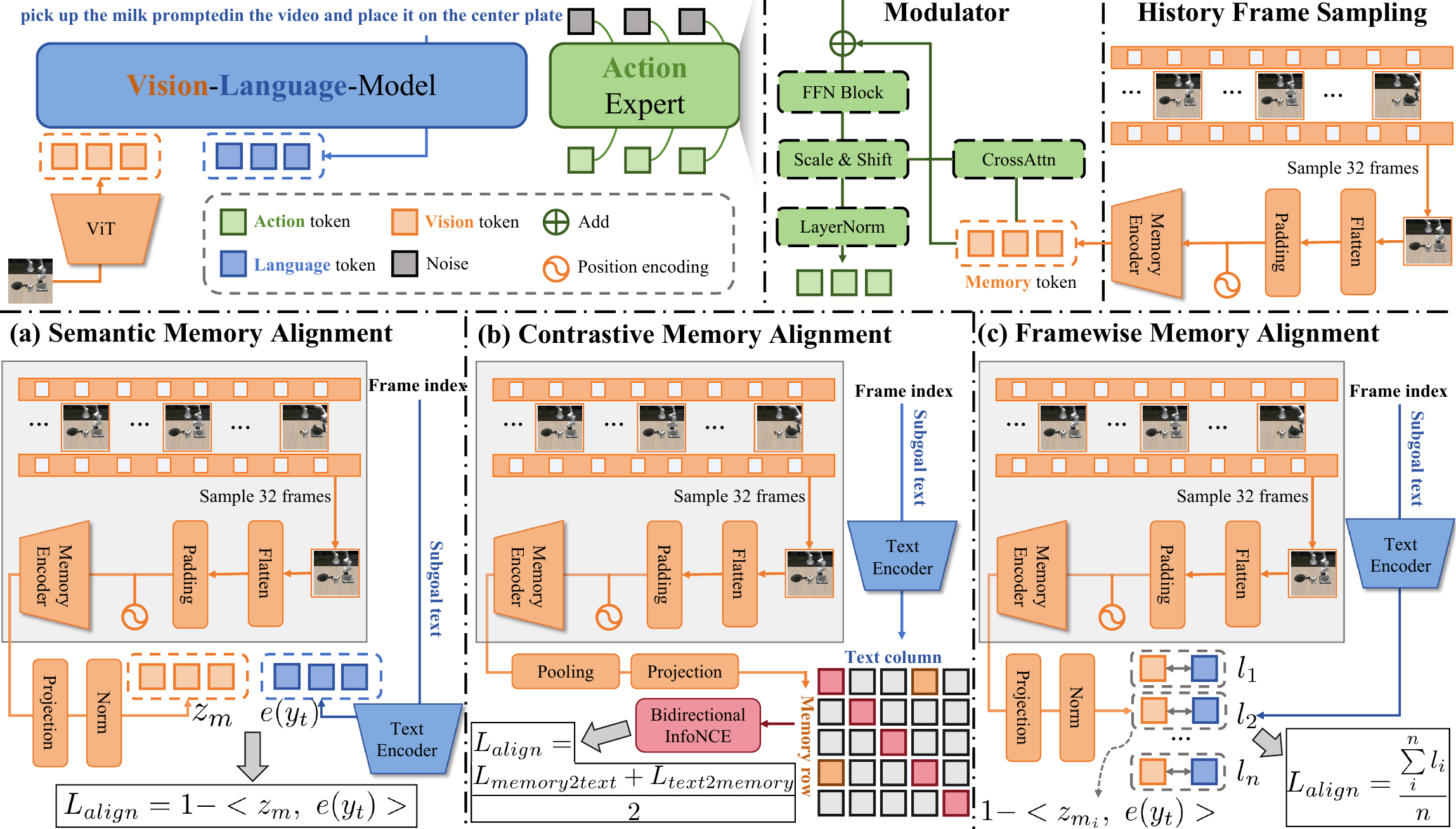}
\caption{
  Overview of memory alignment from annotations. MEMOBench checkpoints provide subgoal text and frame level supervision for Semantic, Contrastive, and Framewise Memory Alignment.
  }
  \label{fig:main}
\end{figure*}

The checkpoint annotations also provide training time supervision for policies with memory augmentation. Figure~\ref{fig:main} summarizes three objectives for the RoboMME $\pi_{0.5}$ memory module~\cite{dai2026robommebenchmarkingunderstandingmemory}: Semantic Memory Alignment uses cosine similarity between pooled memory and text, Contrastive Memory Alignment applies a bidirectional Information Noise-Contrastive Estimation (InfoNCE) loss~\cite{oord2019representationlearningcontrastivepredictive} to paired memory and text representations, and Framewise Memory Alignment matches sampled history frames with temporally aligned checkpoint text. In all cases, an alignment loss is added as a training regularizer:
\[
\mathcal{L}_{\mathrm{total}}
=
\mathcal{L}_{\mathrm{action}}
+\lambda \mathcal{L}_{\mathrm{align}}.
\]
Here, $\mathcal{L}_{\mathrm{align}}$ is the Semantic, Contrastive, or Framewise Memory Alignment objective, and $\lambda$ is chosen per objective. All objectives are used only during training. Appendix~\ref{sec:appendix_alignment_objectives} gives the full definitions.

\section{Evaluation Metrics}
\label{sec:metrics}
MEMOBench reports task success and three process level memory metrics. \textbf{Task Success Rate} (SR) follows the final BDDL goal:
\[
    \mathrm{SR}(\tau)=\frac{1}{N}\sum_{i=1}^{N}\mathbbm{1}[d_i\text{ succeeds}],
\]
where $d_i$ succeeds only when all goal predicates of task $\tau$ are satisfied at termination. SR measures end to end manipulation performance but does not localize memory failures.

The process metrics evaluate executable checkpoints from Section~\ref{sec:benchmark}. Each checkpoint is labeled Storage ($\texttt{S}$), Update ($\texttt{U}$), or Compression ($\texttt{C}$), and its predicate is evaluated on simulator state. For operation $o\in\{\texttt{S},\texttt{U},\texttt{C}\}$ with checkpoint set $\mathcal{C}^{o}_{\tau}$, we compute
\[
    r_o(\tau)=
    \frac{1}{N}\sum_{i=1}^{N}
    \frac{1}{|\mathcal{C}^{o}_{\tau}|}
    \sum_{c\in\mathcal{C}^{o}_{\tau}} v_i(c),
\]
where $v_i(c)=1$ when episode $i$ satisfies the checkpoint predicate and $0$ otherwise. We report $r_{\texttt{S}}$ as \textbf{Memory Storage Rate} (MSR), $r_{\texttt{U}}$ as \textbf{Memory Update Rate} (MUR), and $r_{\texttt{C}}$ as \textbf{Memory Compression Rate} (MCR). Tasks without an operation are marked N/A and excluded from its macro average.

Together, SR and process metrics separate task completion from memory fidelity. Low SR with high process scores suggests action execution as the likely bottleneck, while selective drops in MUR or MCR indicate failures in revising stale state or consolidating history. Appendix~\ref{sec:appendix_metrics} gives the full notation.

\section{Experiments}
\label{sec:experiments}
\noindent The experiments ask whether current VLA policies solve MEMOBench, which memory operations fail, and whether checkpoint annotations supervise memory alignment.

\subsection{Experimental Setup}

We evaluate each policy on all 30 MEMOBench tasks with 1{,}500 training demonstrations and 1{,}500 held out simulator rollouts. Each task has 50 demonstrations and 50 evaluation episodes. Rollouts use seed base 7, 10 stabilization steps, a 1{,}200 step horizon, and replan intervals $k \in \{5,20,35,50\}$, where the evaluator executes the first $k$ actions from each 50 action prediction.

We report task success rate (SR), memory storage rate (MSR), memory update rate (MUR), and memory compression rate (MCR), grouped by temporal, spatial, object, and procedural memory with macro averages over applicable memory types.

We compare standard policies, $\pi_0$~\cite{black2026pi0visionlanguageactionflowmodel}, $\pi_{0.5}$~\cite{intelligence2025pi05visionlanguageactionmodelopenworld}, and OpenVLA-OFT~\cite{kim2025finetuningvisionlanguageactionmodelsoptimizing}, with MemoryVLA~\cite{shi2026memoryvlaperceptualcognitivememoryvisionlanguageaction}, HiF-VLA~\cite{lin2026hifvlahindsightinsightforesight}, and the RoboMME $\pi_{0.5}$ memory module~\cite{dai2026robommebenchmarkingunderstandingmemory}. All rows are fine tuned on MEMOBench. The alignment study uses the same backbone and checkpoint rule for every variant. Appendix~\ref{sec:appendix_reproducibility} gives reproducibility settings.

\subsection{Main Results}

\begin{table*}[t]
\centering
\renewcommand{\arraystretch}{1.14}
\setlength{\tabcolsep}{4.5pt}
\resizebox{\textwidth}{!}{
\begin{tabular}{@{}cl cccc cccc cccc cccc cccc@{}}
\toprule
& & \multicolumn{4}{c}{\textbf{Temporal Mem.}} & \multicolumn{4}{c}{\textbf{Spatial Mem.}} & \multicolumn{4}{c}{\textbf{Object Mem.}} & \multicolumn{4}{c}{\textbf{Procedural Mem.}} & \multicolumn{4}{c}{\textbf{Average}} \\
\cmidrule(lr){3-6} \cmidrule(lr){7-10} \cmidrule(lr){11-14} \cmidrule(lr){15-18} \cmidrule(lr){19-22}
& \textbf{Model}
& SR & MSR & MUR & MCR
& SR & MSR & MUR & MCR
& SR & MSR & MUR & MCR
& SR & MSR & MUR & MCR
& \textbf{Avg. SR} & \textbf{Avg. MSR} & \textbf{Avg. MUR} & \textbf{Avg. MCR} \\
\midrule
\multirow{6}{*}{\rotatebox[origin=c]{90}{\small\textit{step\,=\,5}}}
& $\pi_0$~\cite{black2026pi0visionlanguageactionflowmodel}
& \underline{4.4} & \textbf{94.0} & 12.4 & \underline{4.0} & 0.2 & 0.4 & 0.4 & 0.3 & 27.6 & 27.6 & -- & -- & 12.4 & 32.4 & 12.0 & -- & 11.2 & 38.6 & 8.3 & 2.1 \\
& $\pi_{0.5}$~\cite{intelligence2025pi05visionlanguageactionmodelopenworld}
& 2.2 & 67.6 & 18.4 & 2.0 & 0.0 & 6.4 & 0.3 & \underline{6.4} & \textbf{99.6} & \textbf{99.6} & -- & -- & \underline{19.6} & \underline{46.0} & \underline{22.0} & -- & \underline{30.4} & \underline{54.9} & 13.5 & 4.2 \\
& OpenVLA-OFT~\cite{kim2025finetuningvisionlanguageactionmodelsoptimizing}
& 0.4 & 5.6 & 0.5 & 1.0 & 0.0 & 0.5 & 1.6 & 0.3 & 33.2 & 33.2 & -- & -- & 6.4 & 9.6 & 15.3 & -- & 10.0 & 12.2 & 5.8 & 0.6 \\
\cmidrule(l){2-22}
& MemoryVLA$^\dagger$~\cite{shi2026memoryvlaperceptualcognitivememoryvisionlanguageaction}
& 0.2 & 87.8 & \textbf{38.0} & 0.0 & \textbf{22.4} & \textbf{23.6} & \textbf{24.1} & \textbf{22.3} & 2.4 & 2.4 & -- & -- & \textbf{43.6} & \textbf{58.4} & \textbf{45.7} & -- & 17.2 & 43.1 & \textbf{35.9} & \textbf{11.1} \\
& HiF-VLA$^\dagger$~\cite{lin2026hifvlahindsightinsightforesight}
& 0.2 & 1.8 & 0.0 & 0.0 & 0.0 & 0.3 & 0.0 & 0.1 & \underline{79.6} & \underline{79.6} & -- & -- & 4.0 & 10.4 & 14.3 & -- & 21.0 & 23.0 & 4.8 & 0.1 \\
& $\pi_{0.5}$ + RoboMME Memory Module$^\dagger$~\cite{dai2026robommebenchmarkingunderstandingmemory}
& \textbf{9.8} & \underline{92.0} & \underline{32.2} & \textbf{10.0} & \underline{2.0} & \underline{8.4} & \underline{1.7} & 5.6 & \textbf{99.6} & \textbf{99.6} & -- & -- & 16.0 & 32.4 & 15.7 & -- & \textbf{31.9} & \textbf{58.1} & \underline{16.5} & \underline{7.8} \\
\midrule
\multirow{6}{*}{\rotatebox[origin=c]{90}{\small\textit{step\,=\,20}}}
& $\pi_0$~\cite{black2026pi0visionlanguageactionflowmodel}
& 0.4 & 58.2 & 3.1 & \underline{2.0} & 0.0 & 5.7 & 0.0 & 5.9 & 13.6 & 13.6 & -- & -- & 1.6 & 10.4 & 16.3 & -- & 3.9 & 22.0 & 6.5 & 3.9 \\
& $\pi_{0.5}$~\cite{intelligence2025pi05visionlanguageactionmodelopenworld}
& 0.2 & 40.0 & 25.5 & 0.0 & 1.8 & 10.3 & 2.3 & \underline{8.4} & \textbf{99.6} & \textbf{99.6} & -- & -- & 3.6 & 8.0 & 4.0 & -- & 26.3 & 39.5 & 10.6 & 4.2 \\
& OpenVLA-OFT~\cite{kim2025finetuningvisionlanguageactionmodelsoptimizing}
& 0.4 & 79.6 & 40.4 & 1.0 & 0.4 & 0.5 & 0.5 & 0.3 & \underline{35.2} & \underline{35.2} & -- & -- & \underline{12.4} & \underline{19.6} & \underline{31.3} & -- & 12.1 & 33.7 & 24.1 & 0.6 \\
\cmidrule(l){2-22}
& MemoryVLA$^\dagger$~\cite{shi2026memoryvlaperceptualcognitivememoryvisionlanguageaction}
& 0.2 & \underline{79.8} & \underline{44.0} & 0.0 & \textbf{28.4} & \textbf{41.6} & \textbf{42.1} & \textbf{28.3} & 32.4 & 32.4 & -- & -- & \textbf{55.6} & \textbf{66.4} & \textbf{63.7} & -- & \underline{29.2} & \textbf{55.1} & \textbf{49.9} & \textbf{14.1} \\
& HiF-VLA$^\dagger$~\cite{lin2026hifvlahindsightinsightforesight}
& \textbf{8.6} & 64.0 & \textbf{67.6} & 0.0 & 0.2 & 0.3 & 1.6 & 0.3 & 33.6 & 33.6 & -- & -- & 8.0 & 14.4 & 22.0 & -- & 12.6 & 28.1 & \underline{30.4} & 0.1 \\
& $\pi_{0.5}$ + RoboMME Memory Module$^\dagger$~\cite{dai2026robommebenchmarkingunderstandingmemory}
& \underline{7.6} & \textbf{90.0} & 31.5 & \textbf{8.0} & \underline{4.0} & \underline{12.4} & \underline{3.7} & 7.6 & \textbf{99.6} & \textbf{99.6} & -- & -- & 8.4 & 10.4 & 8.0 & -- & \textbf{29.9} & \underline{53.1} & 14.4 & \underline{7.8} \\
\midrule
\multirow{6}{*}{\rotatebox[origin=c]{90}{\small\textit{step\,=\,35}}}
& $\pi_0$~\cite{black2026pi0visionlanguageactionflowmodel}
& 0.4 & 51.6 & 4.0 & 0.0 & \underline{1.6} & \textbf{24.4} & 2.4 & \textbf{17.6} & 35.6 & 35.6 & -- & -- & 0.4 & 2.0 & 0.3 & -- & 9.5 & 28.4 & 2.2 & \underline{8.8} \\
& $\pi_{0.5}$~\cite{intelligence2025pi05visionlanguageactionmodelopenworld}
& 0.4 & 40.4 & 20.0 & 0.0 & 0.2 & 1.6 & 0.1 & 2.0 & \underline{98.8} & \underline{98.8} & -- & -- & 0.0 & 0.4 & 0.0 & -- & \underline{24.9} & 35.3 & 6.7 & 1.0 \\
& OpenVLA-OFT~\cite{kim2025finetuningvisionlanguageactionmodelsoptimizing}
& 0.0 & 71.6 & \underline{42.4} & 1.0 & 0.0 & 0.5 & 0.5 & 0.3 & 13.2 & 13.2 & -- & -- & \underline{18.4} & \underline{21.6} & \underline{29.3} & -- & 7.9 & 26.7 & 24.1 & 0.6 \\
\cmidrule(l){2-22}
& MemoryVLA$^\dagger$~\cite{shi2026memoryvlaperceptualcognitivememoryvisionlanguageaction}
& 0.2 & \underline{77.8} & 42.0 & 0.0 & \textbf{6.4} & \underline{5.6} & \textbf{6.1} & \underline{6.3} & 6.4 & 6.4 & -- & -- & \textbf{47.6} & \textbf{80.4} & \textbf{59.7} & -- & 15.2 & \underline{42.6} & \textbf{35.9} & 3.1 \\
& HiF-VLA$^\dagger$~\cite{lin2026hifvlahindsightinsightforesight}
& \textbf{9.6} & 40.0 & \textbf{61.6} & \textbf{20.0} & 0.2 & 0.3 & \underline{3.6} & 0.3 & 23.6 & 23.6 & -- & -- & 0.4 & 10.4 & 8.0 & -- & 8.5 & 18.6 & \underline{24.4} & \textbf{10.1} \\
& $\pi_{0.5}$ + RoboMME Memory Module$^\dagger$~\cite{dai2026robommebenchmarkingunderstandingmemory}
& \underline{5.6} & \textbf{86.0} & 27.5 & \underline{6.0} & 0.2 & 4.4 & 0.1 & 3.6 & \textbf{99.2} & \textbf{99.2} & -- & -- & 6.4 & 6.4 & 10.0 & -- & \textbf{27.9} & \textbf{49.0} & 12.5 & 4.8 \\
\midrule
\multirow{6}{*}{\rotatebox[origin=c]{90}{\small\textit{step\,=\,50}}}
& $\pi_0$~\cite{black2026pi0visionlanguageactionflowmodel}
& 0.0 & 72.0 & 0.7 & 0.0 & 0.4 & \textbf{24.3} & 1.6 & \underline{9.6} & 10.4 & 10.4 & -- & -- & \underline{4.4} & 6.4 & 5.7 & -- & 3.8 & 28.3 & 2.7 & 4.8 \\
& $\pi_{0.5}$~\cite{intelligence2025pi05visionlanguageactionmodelopenworld}
& 2.2 & 35.6 & 27.3 & 0.0 & \underline{4.4} & \underline{16.3} & \underline{3.6} & \textbf{12.4} & \underline{76.4} & \underline{76.4} & -- & -- & 0.0 & 0.0 & 1.7 & -- & \underline{20.8} & 32.1 & 10.8 & \underline{6.2} \\
& OpenVLA-OFT~\cite{kim2025finetuningvisionlanguageactionmodelsoptimizing}
& \textbf{10.0} & 75.6 & \underline{56.4} & \underline{10.0} & 0.4 & 0.5 & 0.5 & 0.3 & 31.2 & 31.2 & -- & -- & 2.4 & 5.6 & \underline{19.3} & -- & 11.0 & 28.2 & 25.4 & 5.1 \\
\cmidrule(l){2-22}
& MemoryVLA$^\dagger$~\cite{shi2026memoryvlaperceptualcognitivememoryvisionlanguageaction}
& 0.4 & \textbf{92.4} & 18.4 & 0.0 & \textbf{10.2} & 13.7 & \textbf{14.3} & \underline{9.6} & 9.6 & 9.6 & -- & -- & \textbf{47.6} & \textbf{52.4} & \textbf{56.3} & -- & 17.0 & \textbf{42.0} & \underline{29.7} & 4.8 \\
& HiF-VLA$^\dagger$~\cite{lin2026hifvlahindsightinsightforesight}
& 1.6 & 1.6 & \textbf{91.6} & \textbf{64.0} & 0.0 & 0.3 & 0.0 & 0.1 & 8.4 & 8.4 & -- & -- & 4.0 & \underline{13.6} & 10.3 & -- & 3.5 & 6.0 & \textbf{34.0} & \textbf{32.1} \\
& $\pi_{0.5}$ + RoboMME Memory Module$^\dagger$~\cite{dai2026robommebenchmarkingunderstandingmemory}
& \underline{3.6} & \underline{80.0} & 19.5 & 4.0 & 2.0 & 4.4 & 1.7 & 5.6 & \textbf{79.6} & \textbf{79.6} & -- & -- & 2.0 & 2.4 & 1.7 & -- & \textbf{21.8} & \underline{41.6} & 7.6 & 4.8 \\
\bottomrule
\end{tabular}}
\caption{
Main MEMOBench results under replan steps 5, 20, 35, and 50. Avg. columns macro average over applicable memory types. Bold and underline mark the best and second best values within each replan block. Dashes are excluded from averages, and $^\dagger$ marks explicit memory mechanisms.
}
\label{tab:main_results}
\end{table*}

\begin{table*}[t]
\centering
\renewcommand{\arraystretch}{1.14}
\setlength{\tabcolsep}{4.5pt}
\resizebox{\textwidth}{!}{
\begin{tabular}{@{}cl cccc cccc cccc cccc cccc@{}}
\toprule
& & \multicolumn{4}{c}{\textbf{Temporal Mem.}} & \multicolumn{4}{c}{\textbf{Spatial Mem.}} & \multicolumn{4}{c}{\textbf{Object Mem.}} & \multicolumn{4}{c}{\textbf{Procedural Mem.}} & \multicolumn{4}{c}{\textbf{Average}} \\
\cmidrule(lr){3-6} \cmidrule(lr){7-10} \cmidrule(lr){11-14} \cmidrule(lr){15-18} \cmidrule(lr){19-22}
& \textbf{Method}
& SR & MSR & MUR & MCR
& SR & MSR & MUR & MCR
& SR & MSR & MUR & MCR
& SR & MSR & MUR & MCR
& \textbf{Avg. SR} & \textbf{Avg. MSR} & \textbf{Avg. MUR} & \textbf{Avg. MCR} \\
\midrule
\multirow{4}{*}{\rotatebox[origin=c]{90}{\small\textit{step\,=\,5}}}
& Contrastive Memory Alignment
& \textbf{14.4} & 91.8 & \textbf{39.6} & \textbf{14.0} & \underline{1.4} & \underline{7.6} & \underline{1.6} & \textbf{8.3} & \underline{99.2} & \underline{99.2} & -- & -- & \textbf{23.6} & \textbf{40.4} & \textbf{23.0} & -- & \textbf{34.7} & \underline{59.8} & \textbf{21.4} & \textbf{11.1} \\
& Semantic Memory Alignment
& \underline{12.4} & \underline{97.8} & \underline{36.5} & \underline{12.0} & 0.2 & 5.6 & 0.5 & 4.3 & \underline{99.2} & \underline{99.2} & -- & -- & 15.6 & \textbf{40.4} & 15.0 & -- & 31.9 & \textbf{60.8} & \underline{17.4} & \underline{8.1} \\
& Framewise Memory Alignment
& 10.4 & \textbf{99.4} & 25.6 & 10.0 & \underline{1.4} & 3.6 & \underline{1.6} & 4.3 & \underline{99.2} & \underline{99.2} & -- & -- & \underline{19.6} & \underline{34.4} & \underline{19.0} & -- & \underline{32.7} & 59.2 & 15.4 & 7.1 \\
& Memory module
& 9.8 & 92.0 & 32.2 & 10.0 & \textbf{2.0} & \textbf{8.4} & \textbf{1.7} & \underline{5.6} & \textbf{99.6} & \textbf{99.6} & -- & -- & 16.0 & 32.4 & 15.7 & -- & 31.9 & 58.1 & 16.5 & 7.8 \\
\midrule
\multirow{4}{*}{\rotatebox[origin=c]{90}{\small\textit{step\,=\,20}}}
& Contrastive Memory Alignment
& \textbf{12.4} & \textbf{97.8} & 25.6 & \textbf{12.0} & 1.4 & 7.6 & 1.6 & \underline{6.3} & \underline{99.2} & \underline{99.2} & -- & -- & \underline{5.6} & \textbf{14.4} & 5.0 & -- & \underline{29.7} & \textbf{54.8} & 10.7 & \textbf{9.1} \\
& Semantic Memory Alignment
& \underline{8.4} & 89.8 & \textbf{32.5} & \underline{8.0} & 1.4 & 7.6 & 1.6 & 4.3 & \underline{99.2} & \underline{99.2} & -- & -- & 1.6 & 6.4 & \underline{7.0} & -- & 27.7 & 50.8 & \underline{13.7} & 6.1 \\
& Framewise Memory Alignment
& \underline{8.4} & \underline{93.8} & 23.6 & \underline{8.0} & \underline{3.4} & \underline{9.6} & \underline{3.6} & 4.3 & \underline{99.2} & \underline{99.2} & -- & -- & \underline{5.6} & 8.4 & 5.0 & -- & 29.2 & 52.8 & 10.7 & 6.1 \\
& Memory module
& 7.6 & 90.0 & \underline{31.5} & \underline{8.0} & \textbf{4.0} & \textbf{12.4} & \textbf{3.7} & \textbf{7.6} & \textbf{99.6} & \textbf{99.6} & -- & -- & \textbf{8.4} & \underline{10.4} & \textbf{8.0} & -- & \textbf{29.9} & \underline{53.1} & \textbf{14.4} & \underline{7.8} \\
\midrule
\multirow{4}{*}{\rotatebox[origin=c]{90}{\small\textit{step\,=\,35}}}
& Contrastive Memory Alignment
& 6.4 & 79.8 & 18.5 & 6.0 & \underline{0.2} & \textbf{5.6} & \underline{0.5} & \underline{4.3} & \textbf{99.2} & \textbf{99.2} & -- & -- & 1.6 & 3.2 & 2.0 & -- & 26.9 & 47.0 & 7.0 & 5.1 \\
& Semantic Memory Alignment
& \textbf{10.4} & \underline{87.8} & \textbf{32.5} & \textbf{10.0} & \textbf{1.4} & \textbf{5.6} & \textbf{1.6} & 2.3 & \textbf{99.2} & \textbf{99.2} & -- & -- & 0.0 & 0.4 & 1.0 & -- & \underline{27.8} & 48.3 & \underline{11.7} & \underline{6.1} \\
& Framewise Memory Alignment
& \underline{8.4} & \textbf{89.8} & 25.6 & \underline{8.0} & \underline{0.2} & 3.6 & \underline{0.5} & \textbf{6.3} & \textbf{99.2} & \textbf{99.2} & -- & -- & \underline{3.6} & \underline{4.4} & \underline{5.0} & -- & \textbf{27.9} & \textbf{49.3} & 10.4 & \textbf{7.1} \\
& Memory module
& 5.6 & 86.0 & \underline{27.5} & 6.0 & \underline{0.2} & \underline{4.4} & 0.1 & 3.6 & \textbf{99.2} & \textbf{99.2} & -- & -- & \textbf{6.4} & \textbf{6.4} & \textbf{10.0} & -- & \textbf{27.9} & \underline{49.0} & \textbf{12.5} & 4.8 \\
\midrule
\multirow{4}{*}{\rotatebox[origin=c]{90}{\small\textit{step\,=\,50}}}
& Contrastive Memory Alignment
& 2.4 & 59.8 & 9.6 & 2.0 & 1.4 & \underline{9.6} & 1.6 & 4.3 & \textbf{99.2} & \textbf{99.2} & -- & -- & \underline{0.0} & \underline{0.4} & \underline{1.0} & -- & \underline{25.8} & 42.3 & 4.1 & 3.1 \\
& Semantic Memory Alignment
& \textbf{6.4} & \underline{81.8} & \textbf{26.5} & \textbf{6.0} & \textbf{3.4} & \textbf{13.6} & \textbf{3.6} & \textbf{6.3} & \underline{87.6} & \underline{87.6} & -- & -- & \underline{0.0} & 0.0 & 0.7 & -- & 24.4 & \underline{45.8} & \textbf{10.3} & \textbf{6.1} \\
& Framewise Memory Alignment
& \textbf{6.4} & \textbf{87.8} & \underline{19.6} & \textbf{6.0} & \textbf{3.4} & 7.6 & \textbf{3.6} & \textbf{6.3} & \textbf{99.2} & \textbf{99.2} & -- & -- & \underline{0.0} & \underline{0.4} & \underline{1.0} & -- & \textbf{27.3} & \textbf{48.8} & \underline{8.1} & \textbf{6.1} \\
& Memory module
& \underline{3.6} & 80.0 & 19.5 & \underline{4.0} & \underline{2.0} & 4.4 & \underline{1.7} & \underline{5.6} & 79.6 & 79.6 & -- & -- & \textbf{2.0} & \textbf{2.4} & \textbf{1.7} & -- & 21.8 & 41.6 & 7.6 & \underline{4.8} \\
\bottomrule
\end{tabular}}
\caption{
Memory alignment results for the RoboMME $\pi_{0.5}$ memory module~\cite{dai2026robommebenchmarkingunderstandingmemory}. Alignment variants add auxiliary training losses, while Memory module is the unaligned baseline. Bold and underline mark the best and second best values within each replan block. Dashes are excluded from averages.
}
\label{tab:memory_alignment_results}
\end{table*}

Table~\ref{tab:main_results} shows that current VLA policies are far from saturating MEMOBench. The strongest setting before alignment is $\pi_{0.5}$ with the memory module at replan step 5, reaching 31.9\% average SR, close to standard $\pi_{0.5}$ at 30.4\%. This average masks severe imbalance: standard $\pi_{0.5}$ reaches 99.6\% object SR but only 2.2\% temporal, 0.0\% spatial, and 19.6\% procedural SR, showing that stable object references are easier than changing states, event history, or procedural constraints.

\begin{figure}[t]
\centering
\includegraphics[width=\linewidth]{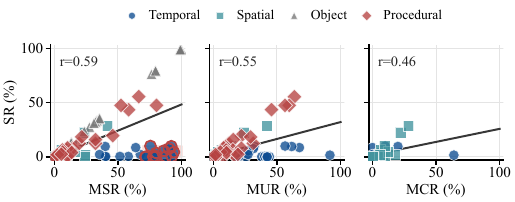}
\caption{Metric correlations from Table~\ref{tab:main_results}. Each point is one model, replan step, and memory type entry. Shading marks temporal cases with high MSR but low SR. Pearson correlations use all applicable entries for each process metric.}
\label{fig:metric_correlation_analysis}
\end{figure}

The process metrics reveal failures that SR alone hides. $\pi_0$ at step 5 reaches 94.0\% temporal MSR but only 4.4\% temporal SR, and $\pi_{0.5}$ with the memory module at step 20 reaches 90.0\% temporal MSR but only 7.6\% temporal SR. These gaps show that policies can store initial information while failing to update or consolidate it for action. Figure~\ref{fig:metric_correlation_analysis} visualizes the same pattern across model, replan, and memory type entries. Across replan settings, the $\pi_{0.5}$ memory module has much higher average MSR than average MUR and MCR, indicating that the bottleneck appears after initial storage.

Explicit memory mechanisms help different operations, but none dominates. MemoryVLA has the strongest update result, with 49.9\% average MUR and 55.6\% procedural SR at step 20, while HiF-VLA reaches the highest MCR at step 50, mainly from temporal memory, but only 3.5\% average SR. $\pi_{0.5}$ with the memory module gives the best average SR under every replan setting, despite low update and compression metrics. These contrasts motivate reporting task and process metrics together. Appendix~\ref{sec:appendix_memer} reports MemER~\cite{sridhar2025memerscalingmemoryrobot} as an additional reference row with privileged checkpoint supervision. It reaches 56.4\% average SR yet still leaves spatial update and procedural memory far from solved.

Replanning frequency also matters. $\pi_{0.5}$ with the memory module drops from 31.9\% average SR at step 5 to 21.8\% at step 50, while standard $\pi_{0.5}$ drops from 30.4\% to 20.8\%. MemoryVLA is the main exception, peaking at step 20 through stronger spatial and procedural performance. Shorter replanning therefore helps memory dependent execution, but it does not remove the need for reliable update and compression.

\subsection{Error Analysis with Failure Mode Taxonomy}

\begin{figure}[t]
\centering
\includegraphics[width=\linewidth]{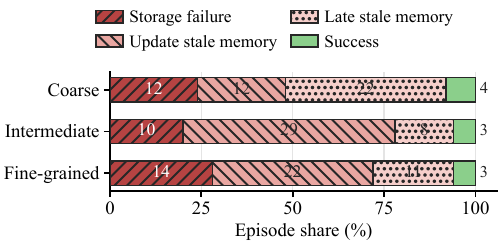}
\caption{Failure taxonomy for a representative temporal task. Bars summarize coarse, intermediate, and fine-grained 50 episode rollout logs.}
\label{fig:failure_taxonomy_analysis}
\end{figure}

Figure~\ref{fig:failure_taxonomy_analysis} localizes temporal failures to stale memory after state changes. Across 150 rollouts of a three checkpoint stove task, only 10 succeed. Among the 140 failures, 36 fail at initial storage, 63 fail from stale memory at the first update checkpoint, and 41 reach the later checkpoint with stale memory. Temporal failures are therefore concentrated in update and stale state persistence rather than only initial encoding.

This taxonomy explains why temporal SR remains low even when storage metrics are high. A policy may store that the stove was turned on, but then act as if the old state remains valid after the stove is turned off. The late stale memory category shows that some episodes pass earlier checkpoints but still fail when the final action depends on the revised state. Thus temporal memory failures mainly arise from revising and using state changes, not from missing the first observation.

\subsection{Oracle Memory Decomposition}

\begin{table}[t]
\centering
\small
\renewcommand{\arraystretch}{1.14}
\setlength{\tabcolsep}{5pt}
\begin{tabular}{@{}l cccc@{}}
\toprule
\textbf{Memory type} & \textbf{SR} & \textbf{MSR} & \textbf{MUR} & \textbf{MCR} \\
\midrule
Temporal & 2.2/54.9 & 67.6/68.0 & 18.4/54.2 & 2.0/53.0 \\
Spatial & 0.0/45.8 & 6.4/77.9 & 0.3/84.1 & 6.4/60.7 \\
Object & 99.6/96.8 & 99.6/96.8 & -- & -- \\
Procedural & 19.6/10.0 & 46.0/13.2 & 22.0/20.3 & -- \\
\midrule
\rowcolor{gray!8}
\textbf{Average} & \textbf{30.4/51.9} & \textbf{54.9/64.0} & \textbf{13.5/52.9} & \textbf{4.2/56.8} \\
\bottomrule
\end{tabular}
\caption{
Oracle memory decomposition at replan step 5. Each cell reports standard $\pi_{0.5}$ and, after the slash, the same backbone fine tuned and evaluated with ground truth memory text appended to the prompt. Dashes mark memory types without the corresponding operation.
}
\label{tab:oracle_memory}
\end{table}

A failed checkpoint can reflect either missing memory content or perception, control, and execution failure. To bound the memory attributable share, we fine tune the standard $\pi_{0.5}$ backbone with ground truth memory text rendered from realized checkpoint annotations appended to the prompt at each replan step, using the same demonstrations, architecture, and training recipe, and evaluate all 30 tasks with 50 rollouts at replan step 5. Table~\ref{tab:oracle_memory} shows that the largest recoveries occur exactly where MEMOBench claims memory failure. Temporal and spatial MUR rise from 18.4\% to 54.2\% and from 0.3\% to 84.1\%, MCR from 2.0\% to 53.0\% and from 6.4\% to 60.7\%, and SR from 2.2\% to 54.9\% and from 0.0\% to 45.8\%, while object memory remains near ceiling and procedural scores do not improve. This selectivity argues against a generic prompt or training effect and supports a sandwich reading of the process metrics. The standard score mixes memory and manipulation, the oracle score estimates the failure that perfect memory content would recover, and the remaining gap to 100\% is the manipulation and utilization floor. On average, oracle memory closes 45.5\% of the MUR gap and 54.9\% of the MCR gap, but only 20.2\% of the MSR gap, consistent with Storage checkpoints being largely satisfiable while the stored information is still visible.

\subsection{Shuffled Memory Control}

\begin{figure}[t]
\centering
\includegraphics[width=\linewidth]{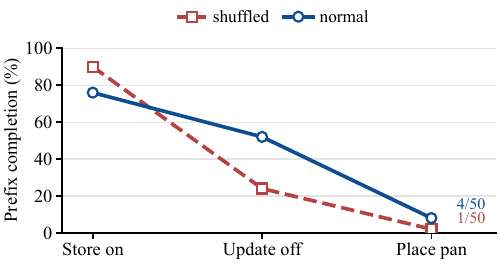}
\caption{Shuffled memory control on the temporal task from Figure~\ref{fig:failure_taxonomy_analysis}. Lines show prefix checkpoint completion, and final labels show successful rollouts.}
\label{fig:shuffled_memory_control}
\end{figure}

Figure~\ref{fig:shuffled_memory_control} tests whether ordered subgoal memory matters for the temporal task. We train the same setting as the coarse annotation ablation, but shuffle memory subgoals during training while keeping the task, rollout protocol, and checkpoint predicates fixed. The shuffled control reaches 90\% completion at the first storage checkpoint, but falls to 24\% after the stove state update and 2\% final success. Normal subgoal order reaches 76\%, 52\%, and 8\% on the same three checkpoints. Thus the control can learn the local initial action, yet loses the advantage when the final action depends on the updated stove state. This supports a narrower conclusion: unordered subgoal exposure is insufficient for the update dependent part of this temporal task, and preserving temporal order provides useful memory supervision.

\subsection{Memory Alignment Results}

Table~\ref{tab:memory_alignment_results} tests whether checkpoint annotations improve memory representations through auxiliary losses. All variants use the same $\pi_{0.5}$ memory module backbone, and checkpoint subgoal text is used only during training. The Memory module row is the controlled baseline without an alignment loss, so differences isolate annotation supervision.

Alignment improves the best setting, but gains vary by replan interval. At step 5, Contrastive Memory Alignment improves average SR from 31.9\% to 34.7\%, temporal SR from 9.8\% to 14.4\%, and procedural SR from 16.0\% to 23.6\%. Across all replan settings, Contrastive and Framewise both average 29.3\% SR, compared with 27.9\% for the unaligned memory module. Framewise is more stable when replanning is sparse, reaching 27.3\% SR at step 50 compared with 21.8\% for the baseline.

The process metrics specialize by objective. Averaging each method's average metric column over the four replan settings, Framewise gives the highest mean MSR, Semantic the highest mean MUR, and Contrastive the highest mean MCR. These differences suggest that checkpoint annotations are useful training signals, while the alignment objective determines which memory operation benefits most.

\subsection{Annotation Granularity Ablation}

\begin{figure}[t]
\centering
\includegraphics[width=\linewidth]{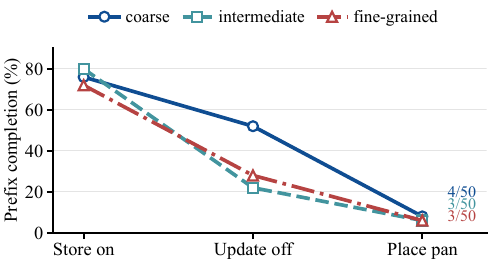}
\caption{Annotation granularity ablation on the temporal task from Figure~\ref{fig:failure_taxonomy_analysis}. Lines show prefix checkpoint completion, and final labels show successful rollouts.}
\label{fig:annotation_granularity_ablation}
\end{figure}

Figure~\ref{fig:annotation_granularity_ablation} shows that more specific checkpoint language does not monotonically improve grounding. Coarse text gives the highest final success rate, 4/50 rollouts, while intermediate and fine-grained text both reach 3/50. Intermediate text has the best initial storage rate at 80\%, but drops sharply at the update checkpoint. Fine-grained text recovers some update performance relative to intermediate text, but does not improve final task success. Useful checkpoint text must therefore preserve the state change that controls later action, not merely describe the task in more detail.

\subsection{Takeaways}

The experiments support three conclusions. First, MEMOBench is not saturated by current VLA policies, even when they use explicit memory modules. Second, initial storage is often easier than update and compression, which explains why high MSR can coexist with low SR. Third, process level annotations are not only diagnostic. They can also supervise memory alignment losses that improve performance without changing the evaluation protocol, and their linguistic granularity affects the learned grounding signal. The oracle memory decomposition further attributes much of the remaining Update and Compression failure to memory itself. Perfect memory content closes 45.5\% of the MUR gap and 54.9\% of the MCR gap, while the residual prices the manipulation and utilization floor. The central challenge for future memory augmented policies is to maintain and consolidate task relevant history, not merely to append more observations to the model input.


\section{Conclusion}
\label{sec:conclusion}
MEMOBench evaluates memory grounded robotic manipulation through task outcomes and process level memory operations. It provides 30 history dependent tasks, 1{,}500 demonstrations, checkpoint language, and executable predicates for Storage, Update, and Compression. Experiments show that standard and memory augmented VLA policies still struggle: they store initial information more reliably than they update stale state or compress history for action. Checkpoint language also provides supervision for memory alignment, letting future systems be compared by the operations they improve rather than only final success.

\section*{Limitations}
While MEMOBench advances memory evaluation for robotic manipulation, several limitations remain and suggest directions for future work.

\paragraph{Simulation Evaluation.}
MEMOBench is constructed entirely in simulation, which enables data collection at large scale and reproducible evaluation but cannot fully capture real world factors such as sensor noise, contact dynamics, and visual domain shift. Validating whether our findings and the proposed process level metrics transfer to physical robots remains an important open direction.

\paragraph{Model Coverage.}
Our main comparison evaluates $\pi_0$, $\pi_{0.5}$, OpenVLA-OFT, MemoryVLA, HiF-VLA, and a $\pi_{0.5}$ memory module, while the controlled alignment study is limited to variants of the same $\pi_{0.5}$ memory module backbone. Other policy architectures, such as state space model architectures, are not evaluated. Extending MEMOBench evaluation to a broader set of model families would strengthen the generalizability of our conclusions.

\paragraph{Annotation Subjectivity.}
The process level memory annotations require human judgment to determine when storage, update, and compression events occur within each demonstration. Although we establish clear guidelines and conduct quality checks, some degree of subjectivity is inherent. A blind relabeling study with three annotators who did not author the released labels shows substantial agreement on the operation labels, with Fleiss' $\kappa=0.799$ and exact three rater agreement on 70 of 84 templates (Appendix~\ref{sec:appendix_annotator_agreement}). The remaining disagreements concentrate on the boundary between Update and Compression. Developing automated or partly automated annotation pipelines could improve scalability and consistency.


\section*{Ethics Statement}
\paragraph{Potential Risks.}
MEMOBench is a simulated benchmark for evaluating memory in robotic manipulation. We do not identify additional risks beyond standard research use of simulated robot benchmarks. The benchmark is intended for diagnostic evaluation and training analysis, not for direct deployment of robot policies in safety critical settings.

\paragraph{Artifacts and Licenses.}
We plan to release the code under the Apache-2.0 license. The dataset, task specifications, and annotations will be released under CC BY 4.0. The work builds on existing robotics and VLA artifacts, including LIBERO, robosuite, MuJoCo, OpenVLA, $\pi_0$, $\pi_{0.5}$, RoboMME, MemoryVLA, HiF-VLA, and SigLIP. We use these artifacts in ways compatible with their terms.

\paragraph{Data Content.}
The dataset is generated in simulation and contains task specifications, robot trajectories, simulator states, rendered observations, and memory checkpoint annotations. It contains no human personal data, faces, speech, real household videos, or offensive text.

\paragraph{Human Annotation.}
The memory checkpoint annotations were created and reviewed by the authors. The blind relabeling study in Appendix~\ref{sec:appendix_annotator_agreement} was conducted by internal lab members who were unaware of the study and did not author the released labels. No external annotators or human subjects were recruited, no payment was involved, and no participant consent procedure or ethics board review was required.

\paragraph{AI Assistance.}
AI assistants were used for sentence polishing. They were not used to generate experimental results, create annotations, or make scientific decisions.

\section*{Acknowledgments}
This work was supported by the National Natural Science Foundation of China under Grant No. U24A20326.

\bibliography{content/ref}

@misc{fang2025sam2actintegratingvisualfoundation,
  author    = {Haoquan Fang and
               Markus Grotz and
               Wilbert Pumacay and
               Yi Ru Wang and
               Dieter Fox and
               Ranjay Krishna and
               Jiafei Duan},
  editor    = {Aarti Singh and
               Maryam Fazel and
               Daniel Hsu and
               Simon Lacoste{-}Julien and
               Felix Berkenkamp and
               Tegan Maharaj and
               Kiri Wagstaff and
               Jerry Zhu},
  title     = {SAM2Act: Integrating Visual Foundation Model with {A} Memory Architecture
               for Robotic Manipulation},
  booktitle = {Forty-second International Conference on Machine Learning, {ICML}
               2025, Vancouver, BC, Canada, July 13-19, 2025},
  series    = {Proceedings of Machine Learning Research},
  publisher = {{PMLR} / OpenReview.net},
  year      = {2025},
  url       = {https://proceedings.mlr.press/v267/fang25c.html},
  bibsource = {dblp computer science bibliography, https://dblp.org}
}

@misc{cherepanov2026memorybenchmarkrobots,
  title         = {Memory, Benchmark \& Robots: A Benchmark for Solving Complex Tasks with Reinforcement Learning},
  author        = {Egor Cherepanov and Nikita Kachaev and Alexey K. Kovalev and Aleksandr I. Panov},
  year          = {2026},
  eprint        = {2502.10550},
  archiveprefix = {arXiv},
  primaryclass  = {cs.LG},
  url           = {https://arxiv.org/abs/2502.10550}
}

@misc{dai2026robommebenchmarkingunderstandingmemory,
  title         = {RoboMME: Benchmarking and Understanding Memory for Robotic Generalist Policies},
  author        = {Yinpei Dai and Hongze Fu and Jayjun Lee and Yuejiang Liu and Haoran Zhang and Jianing Yang and Chelsea Finn and Nima Fazeli and Joyce Chai},
  year          = {2026},
  eprint        = {2603.04639},
  archiveprefix = {arXiv},
  primaryclass  = {cs.RO},
  url           = {https://arxiv.org/abs/2603.04639}
}

@misc{oord2019representationlearningcontrastivepredictive,
  title         = {Representation Learning with Contrastive Predictive Coding},
  author        = {Aaron van den Oord and Yazhe Li and Oriol Vinyals},
  year          = {2019},
  eprint        = {1807.03748},
  archiveprefix = {arXiv},
  primaryclass  = {cs.LG},
  url           = {https://arxiv.org/abs/1807.03748}
}

@misc{liu2023liberobenchmarkingknowledgetransfer,
  author    = {Bo Liu and
               Yifeng Zhu and
               Chongkai Gao and
               Yihao Feng and
               Qiang Liu and
               Yuke Zhu and
               Peter Stone},
  editor    = {Alice Oh and
               Tristan Naumann and
               Amir Globerson and
               Kate Saenko and
               Moritz Hardt and
               Sergey Levine},
  title     = {{LIBERO:} Benchmarking Knowledge Transfer for Lifelong Robot Learning},
  booktitle = {Advances in Neural Information Processing Systems 36: Annual Conference
               on Neural Information Processing Systems 2023, NeurIPS 2023, New Orleans,
               LA, USA, December 10 - 16, 2023},
  year      = {2023},
  url       = {http://papers.nips.cc/paper\_files/paper/2023/hash/8c3c666820ea055a77726d66fc7d447f-Abstract-Datasets\_and\_Benchmarks.html},
  bibsource = {dblp computer science bibliography, https://dblp.org}
}

@misc{mees2022calvinbenchmarklanguageconditionedpolicy,
  author    = {Oier Mees and
               Luk{\'{a}}s Hermann and
               Erick Rosete{-}Beas and
               Wolfram Burgard},
  title     = {{CALVIN:} {A} Benchmark for Language-Conditioned Policy Learning for
               Long-Horizon Robot Manipulation Tasks},
  journal   = {{IEEE} Robotics Autom. Lett.},
  volume    = {7},
  number    = {3},
  pages     = {7327--7334},
  year      = {2022},
  url       = {https://doi.org/10.1109/LRA.2022.3180108},
  doi       = {10.1109/LRA.2022.3180108},
  bibsource = {dblp computer science bibliography, https://dblp.org}
}

@misc{walke2024bridgedatav2datasetrobot,
  author    = {Homer Rich Walke and
               Kevin Black and
               Tony Z. Zhao and
               Quan Vuong and
               Chongyi Zheng and
               Philippe Hansen{-}Estruch and
               Andre Wang He and
               Vivek Myers and
               Moo Jin Kim and
               Max Du and
               Abraham Lee and
               Kuan Fang and
               Chelsea Finn and
               Sergey Levine},
  editor    = {Jie Tan and
               Marc Toussaint and
               Kourosh Darvish},
  title     = {BridgeData {V2:} {A} Dataset for Robot Learning at Scale},
  booktitle = {Conference on Robot Learning, CoRL 2023, 6-9 November 2023, Atlanta,
               GA, {USA}},
  series    = {Proceedings of Machine Learning Research},
  pages     = {1723--1736},
  publisher = {{PMLR}},
  year      = {2023},
  url       = {https://proceedings.mlr.press/v229/walke23a.html},
  bibsource = {dblp computer science bibliography, https://dblp.org}
}

@misc{james2019rlbenchrobotlearningbenchmark,
  author    = {Stephen James and
               Zicong Ma and
               David Rovick Arrojo and
               Andrew J. Davison},
  title     = {RLBench: The Robot Learning Benchmark {\&} Learning Environment},
  journal   = {{IEEE} Robotics Autom. Lett.},
  volume    = {5},
  number    = {2},
  pages     = {3019--3026},
  year      = {2020},
  url       = {https://doi.org/10.1109/LRA.2020.2974707},
  doi       = {10.1109/LRA.2020.2974707},
  bibsource = {dblp computer science bibliography, https://dblp.org}
}

@misc{chung2025rethinkingprogressionmemorystate,
  author    = {Nhat Chung and
               Taisei Hanyu and
               Toan Nguyen and
               Huy Le and
               Frederick Bumgarner and
               Duy Minh Ho Nguyen and
               Khoa Vo and
               Kashu Yamazaki and
               Chase Rainwater and
               Tung Kieu and
               Anh Nguyen and
               Ngan Le},
  editor    = {Sven Koenig and
               Chad Jenkins and
               Matthew E. Taylor},
  title     = {Rethinking Progression of Memory State in Robotic Manipulation: An
               Object-Centric Perspective},
  booktitle = {Fortieth {AAAI} Conference on Artificial Intelligence, Thirty-Eighth
               Conference on Innovative Applications of Artificial Intelligence,
               Sixteenth Symposium on Educational Advances in Artificial Intelligence,
               {AAAI} 2026, Singapore, January 20-27, 2026},
  pages     = {3407--3415},
  publisher = {{AAAI} Press},
  year      = {2026},
  url       = {https://doi.org/10.1609/aaai.v40i5.37337},
  doi       = {10.1609/AAAI.V40I5.37337},
  bibsource = {dblp computer science bibliography, https://dblp.org}
}

@misc{chen2026rmbenchmemorydependentroboticmanipulation,
  title         = {RMBench: Memory-Dependent Robotic Manipulation Benchmark with Insights into Policy Design},
  author        = {Tianxing Chen and Yuran Wang and Mingleyang Li and Yan Qin and Hao Shi and Zixuan Li and Yifan Hu and Yingsheng Zhang and Kaixuan Wang and Yue Chen and Hongcheng Wang and Renjing Xu and Ruihai Wu and Yao Mu and Yaodong Yang and Hao Dong and Ping Luo},
  year          = {2026},
  eprint        = {2603.01229},
  archiveprefix = {arXiv},
  primaryclass  = {cs.RO},
  url           = {https://arxiv.org/abs/2603.01229}
}

@misc{han2025robocerebralargescalebenchmarklonghorizon,
  title         = {RoboCerebra: A Large-scale Benchmark for Long-horizon Robotic Manipulation Evaluation},
  author        = {Songhao Han and Boxiang Qiu and Yue Liao and Siyuan Huang and Chen Gao and Shuicheng Yan and Si Liu},
  year          = {2025},
  eprint        = {2506.06677},
  archiveprefix = {arXiv},
  primaryclass  = {cs.RO},
  url           = {https://arxiv.org/abs/2506.06677}
}

@misc{tao2025maniskill3gpuparallelizedrobotics,
  title         = {ManiSkill3: GPU Parallelized Robotics Simulation and Rendering for Generalizable Embodied AI},
  author        = {Stone Tao and Fanbo Xiang and Arth Shukla and Yuzhe Qin and Xander Hinrichsen and Xiaodi Yuan and Chen Bao and Xinsong Lin and Yulin Liu and Tse-kai Chan and Yuan Gao and Xuanlin Li and Tongzhou Mu and Nan Xiao and Arnav Gurha and Viswesh Nagaswamy Rajesh and Yong Woo Choi and Yen-Ru Chen and Zhiao Huang and Roberto Calandra and Rui Chen and Shan Luo and Hao Su},
  year          = {2025},
  eprint        = {2410.00425},
  archiveprefix = {arXiv},
  primaryclass  = {cs.RO},
  url           = {https://arxiv.org/abs/2410.00425}
}

@incollection{ATKINSON196889,
  author    = {Richard C. Atkinson and
               Richard M. Shiffrin},
  editor    = {Kenneth W. Spence and
               Janet Taylor Spence},
  title     = {Human Memory: {A} Proposed System and its Control Processes},
  booktitle = {Psychology of Learning and Motivation},
  series    = {Psychology of Learning and Motivation},
  pages     = {89--195},
  publisher = {Elsevier},
  year      = {1968},
  url       = {https://doi.org/10.1016/s0079-7421(08)60422-3},
  doi       = {10.1016/S0079-7421(08)60422-3},
  bibsource = {dblp computer science bibliography, https://dblp.org}
}

@misc{black2026pi0visionlanguageactionflowmodel,
  title         = {$\pi_0$: A Vision-Language-Action Flow Model for General Robot Control},
  author        = {Kevin Black and Noah Brown and Danny Driess and Adnan Esmail and Michael Equi and Chelsea Finn and Niccolo Fusai and Lachy Groom and Karol Hausman and Brian Ichter and Szymon Jakubczak and Tim Jones and Liyiming Ke and Sergey Levine and Adrian Li-Bell and Mohith Mothukuri and Suraj Nair and Karl Pertsch and Lucy Xiaoyang Shi and James Tanner and Quan Vuong and Anna Walling and Haohuan Wang and Ury Zhilinsky},
  year          = {2026},
  eprint        = {2410.24164},
  archiveprefix = {arXiv},
  primaryclass  = {cs.LG},
  url           = {https://arxiv.org/abs/2410.24164}
}

@misc{intelligence2025pi05visionlanguageactionmodelopenworld,
  title         = {$\pi_{0.5}$: a Vision-Language-Action Model with Open-World Generalization},
  author        = {Physical Intelligence and Kevin Black and Noah Brown and James Darpinian and Karan Dhabalia and Danny Driess and Adnan Esmail and Michael Equi and Chelsea Finn and Niccolo Fusai and Manuel Y. Galliker and Dibya Ghosh and Lachy Groom and Karol Hausman and Brian Ichter and Szymon Jakubczak and Tim Jones and Liyiming Ke and Devin LeBlanc and Sergey Levine and Adrian Li-Bell and Mohith Mothukuri and Suraj Nair and Karl Pertsch and Allen Z. Ren and Lucy Xiaoyang Shi and Laura Smith and Jost Tobias Springenberg and Kyle Stachowicz and James Tanner and Quan Vuong and Homer Walke and Anna Walling and Haohuan Wang and Lili Yu and Ury Zhilinsky},
  year          = {2025},
  eprint        = {2504.16054},
  archiveprefix = {arXiv},
  primaryclass  = {cs.LG},
  url           = {https://arxiv.org/abs/2504.16054}
}

@misc{kim2024openvlaopensourcevisionlanguageactionmodel,
  author    = {Moo Jin Kim and
               Karl Pertsch and
               Siddharth Karamcheti and
               Ted Xiao and
               Ashwin Balakrishna and
               Suraj Nair and
               Rafael Rafailov and
               Ethan Paul Foster and
               Pannag R. Sanketi and
               Quan Vuong and
               Thomas Kollar and
               Benjamin Burchfiel and
               Russ Tedrake and
               Dorsa Sadigh and
               Sergey Levine and
               Percy Liang and
               Chelsea Finn},
  editor    = {Pulkit Agrawal and
               Oliver Kroemer and
               Wolfram Burgard},
  title     = {OpenVLA: An Open-Source Vision-Language-Action Model},
  booktitle = {Conference on Robot Learning, 6-9 November 2024, Munich, Germany},
  series    = {Proceedings of Machine Learning Research},
  pages     = {2679--2713},
  publisher = {{PMLR}},
  year      = {2024},
  url       = {https://proceedings.mlr.press/v270/kim25c.html},
  bibsource = {dblp computer science bibliography, https://dblp.org}
}

@misc{octomodelteam2024octoopensourcegeneralistrobot,
  author    = {Dibya Ghosh and
               Homer Rich Walke and
               Karl Pertsch and
               Kevin Black and
               Oier Mees and
               Sudeep Dasari and
               Joey Hejna and
               Tobias Kreiman and
               Charles Xu and
               Jianlan Luo and
               You Liang Tan and
               Lawrence Yunliang Chen and
               Quan Vuong and
               Ted Xiao and
               Pannag R. Sanketi and
               Dorsa Sadigh and
               Chelsea Finn and
               Sergey Levine},
  editor    = {Dana Kulic and
               Gentiane Venture and
               Kostas E. Bekris and
               Enrique Coronado},
  title     = {Octo: An Open-Source Generalist Robot Policy},
  booktitle = {Robotics: Science and Systems XX, Delft, The Netherlands, July 15-19,
               2024},
  year      = {2024},
  url       = {https://doi.org/10.15607/RSS.2024.XX.090},
  doi       = {10.15607/RSS.2024.XX.090},
  bibsource = {dblp computer science bibliography, https://dblp.org}
}

@article{nader_fear_2000,
  title     = {Fear memories require protein synthesis in the amygdala for reconsolidation after retrieval},
  volume    = {406},
  issn      = {1476-4687},
  url       = {http://dx.doi.org/10.1038/35021052},
  doi       = {10.1038/35021052},
  number    = {6797},
  journal   = {Nature},
  publisher = {Springer Science and Business Media LLC},
  author    = {Nader,  Karim and Schafe,  Glenn E. and Le Doux,  Joseph E.},
  year      = {2000},
  month     = Aug,
  pages     = {722–726}
}

@article{miller1956magical,
  title     = {The magical number seven,  plus or minus two: Some limits on our capacity for processing information.},
  volume    = {63},
  issn      = {0033-295X},
  url       = {http://dx.doi.org/10.1037/h0043158},
  doi       = {10.1037/h0043158},
  number    = {2},
  journal   = {Psychological Review},
  publisher = {American Psychological Association (APA)},
  author    = {Miller,  George A.},
  year      = {1956},
  month     = Mar,
  pages     = {81–97}
}

@misc{shi2026memoryvlaperceptualcognitivememoryvisionlanguageaction,
  title     = {Memory{VLA}: Perceptual-Cognitive Memory in Vision-Language-Action Models for Robotic Manipulation},
  author    = {Hao Shi and Bin Xie and Yingfei Liu and Lin Sun and Fengrong Liu and Tiancai Wang and Erjin Zhou and Haoqiang Fan and Xiangyu Zhang and Gao Huang},
  booktitle = {The Fourteenth International Conference on Learning Representations},
  year      = {2026},
  url       = {https://openreview.net/forum?id=54U3XHf7qq}
}

@misc{lin2026hifvlahindsightinsightforesight,
  title         = {HiF-VLA: Hindsight, Insight and Foresight through Motion Representation for Vision-Language-Action Models},
  author        = {Minghui Lin and Pengxiang Ding and Shu Wang and Zifeng Zhuang and Yang Liu and Xinyang Tong and Wenxuan Song and Shangke Lyu and Siteng Huang and Donglin Wang},
  year          = {2026},
  eprint        = {2512.09928},
  archiveprefix = {arXiv},
  primaryclass  = {cs.RO},
  url           = {https://arxiv.org/abs/2512.09928}
}

@misc{kim2025finetuningvisionlanguageactionmodelsoptimizing,
  title         = {Fine-Tuning Vision-Language-Action Models: Optimizing Speed and Success},
  author        = {Moo Jin Kim and Chelsea Finn and Percy Liang},
  year          = {2025},
  eprint        = {2502.19645},
  archiveprefix = {arXiv},
  primaryclass  = {cs.RO},
  url           = {https://arxiv.org/abs/2502.19645}
}

@misc{lei2026robomemarenacomprehensivechallengingrobotic,
  title         = {RoboMemArena: A Comprehensive and Challenging Robotic Memory Benchmark},
  author        = {Huashuo Lei and Wenxuan Song and Huarui Zhang and Jieyuan Pei and Jiayi Chen and Haodong Yan and Han Zhao and Pengxiang Ding and Zhipeng Zhang and Lida Huang and Donglin Wang and Yan Wang and Haoang Li},
  year          = {2026},
  eprint        = {2605.10921},
  archiveprefix = {arXiv},
  primaryclass  = {cs.RO},
  url           = {https://arxiv.org/abs/2605.10921}
}

@misc{sridhar2025memerscalingmemoryrobot,
  title         = {MemER: Scaling Up Memory for Robot Control via Experience Retrieval},
  author        = {Ajay Sridhar and Jennifer Pan and Satvik Sharma and Chelsea Finn},
  year          = {2025},
  eprint        = {2510.20328},
  archiveprefix = {arXiv},
  primaryclass  = {cs.RO},
  url           = {https://arxiv.org/abs/2510.20328}
}

@article{liu2018robotintelligencerealworld,
  author   = {LIU, Yunhui and ZHENG, Fan and GUO, Ruibin and WANG, Jiangliu and NIE, Qiang and WANG, Xin and WANG, Zerui},
  title    = {Robot Intelligence for Real World Applications},
  journal  = {Chinese Journal of Electronics},
  volume   = {27},
  number   = {3},
  pages    = {446-458},
  doi      = {https://doi.org/10.1049/cje.2018.03.007},
  url      = {https://ietresearch.onlinelibrary.wiley.com/doi/abs/10.1049/cje.2018.03.007},
  eprint   = {https://ietresearch.onlinelibrary.wiley.com/doi/pdf/10.1049/cje.2018.03.007},
  year     = {2018}
}

@article{zheng2025machinememoryintelligence,
  title    = {Machine Memory Intelligence: Inspired by Human Memory Mechanisms},
  journal  = {Engineering},
  volume   = {55},
  pages    = {24-35},
  year     = {2025},
  issn     = {2095-8099},
  doi      = {https://doi.org/10.1016/j.eng.2025.01.012},
  url      = {https://www.sciencedirect.com/science/article/pii/S2095809925000293},
  author   = {Qinghua Zheng and Huan Liu and Xiaoqing Zhang and Caixia Yan and Xiangyong Cao and Tieliang Gong and Yong-Jin Liu and Bin Shi and Zhen Peng and Xiaocen Fan and Ying Cai and Jun Liu}
}

\appendix
\section{Task Suite Details}
\label{sec:appendix_task_suite}

\begin{table*}[t]
\centering
\footnotesize
\renewcommand{\arraystretch}{1.12}
\setlength{\tabcolsep}{4.5pt}
\begin{tabular}{@{}
>{\raggedright\arraybackslash}p{0.09\textwidth}
>{\raggedright\arraybackslash}p{0.18\textwidth}
>{\raggedright\arraybackslash}p{0.45\textwidth}
>{\centering\arraybackslash}p{0.06\textwidth}
>{\centering\arraybackslash}p{0.06\textwidth}
>{\centering\arraybackslash}p{0.06\textwidth}@{}}
\toprule
\textbf{Type} & \textbf{Family} & \textbf{Task Variants} & \textbf{Ops.} & \textbf{\#Task} & \textbf{\#Ckpt.} \\
\midrule
\multirow{4}{*}{Temporal}
& Drawer checking & Top drawer, middle drawer, all three drawers & S/U/C & 3 & 5 \\
& Appliance tracking & Microwave open/close, microwave open/close plus bowl placement & S/U/C & 2 & 5 \\
& Stove transition & Stove on/off, near shelf variant, two frying pan placement variants & S/U & 4 & 10 \\
& Drawer transfer & Check middle drawer, then place black bowl in top drawer and close it & S/U & 1 & 3 \\
\addlinespace[2pt]
\multirow{2}{*}{Spatial}
& Single drawer placement & Artichoke, broccoli, brussels sprouts, charger, or tomato into top drawer & S/U/C & 5 & 15 \\
& Two drawer placement & Five paired object variants for top and middle drawers & S/U/C & 5 & 30 \\
\addlinespace[2pt]
Object & Referent selection & Milk or butter selected by prior pick, highlighted location, pointed reference, or target bowl & S & 5 & 5 \\
Procedural & Demonstrated routines & Assembly, clear and serve, discard then plate, bowl stacking, buffered object swap & S/U & 5 & 11 \\
\midrule
\rowcolor{gray!8}
\textbf{Total} & \multicolumn{2}{l}{\textbf{All task families}} & \textbf{S/U/C} & \textbf{30} & \textbf{84} \\
\bottomrule
\end{tabular}
\caption{
Task family summary for MEMOBench. Ops. gives the memory operations covered by each family, where S = Storage, U = Update, and C = Compression. \#Ckpt. reports executable checkpoint templates before expanding them across demonstrations.
}
\label{tab:task_suite}
\end{table*}

Table~\ref{tab:task_suite} expands the 30 MEMOBench tasks into 84 executable checkpoint templates. At the default evaluation granularity, these templates contain 35 Storage, 32 Update, and 17 Compression checkpoints. With 50 demonstrations per task, the checkpoint templates yield 4{,}200 demonstration level instances, including 1{,}750 Storage, 1{,}600 Update, and 850 Compression instances. The task suite covers all four memory types, but operation counts are intentionally uneven because object reference tasks mainly test Storage and spatial drawer tasks contribute all three operations.

\begin{table*}[t]
\centering
\small
\renewcommand{\arraystretch}{1.1}
\setlength{\tabcolsep}{5pt}
\begin{tabular}{@{}lrrrrrrrr@{}}
\toprule
\textbf{Family} & \textbf{Tasks} & \textbf{Episodes} & \textbf{Steps} & \textbf{Avg.} & \textbf{Median} & \textbf{P95} & \textbf{Max} & \textbf{Storage} \\
\midrule
Object & 5 & 250 & 121{,}276 & 485.10 & 492.00 & 526.00 & 556 & 44.53 GiB \\
Procedural & 5 & 250 & 250{,}204 & 1{,}000.82 & 876.00 & 1{,}649.50 & 2{,}012 & 91.79 GiB \\
Spatial & 10 & 500 & 357{,}865 & 715.73 & 727.00 & 1{,}101.10 & 1{,}513 & 131.58 GiB \\
Temporal & 10 & 500 & 198{,}099 & 396.20 & 400.50 & 679.05 & 779 & 72.66 GiB \\
\midrule
\rowcolor{gray!8}
\textbf{Total} & \textbf{30} & \textbf{1{,}500} & \textbf{927{,}444} & \textbf{618.30} & \textbf{489.00} & \textbf{1{,}110.10} & \textbf{2{,}012} & \textbf{340.56 GiB} \\
\bottomrule
\end{tabular}
\caption{
Family level statistics for the expert demonstrations. Steps count action or control steps, and storage is measured from the HDF5 task files.
}
\label{tab:family_statistics}
\end{table*}

Table~\ref{tab:family_statistics} gives the scale of the collected demonstrations by memory family. The full dataset contains 927{,}444 action steps and occupies 340.56 GiB. Because each task contributes 50 demonstrations, family differences mainly reflect horizon length rather than sampling imbalance. Procedural tasks have the longest demonstrations, with a mean of 1{,}000.82 steps and a maximum of 2{,}012 steps, because routines such as buffered object swapping require several ordered manipulation phases. Spatial tasks are tied with temporal tasks by count, with 10 tasks and 500 episodes, and they occupy the most storage at 131.58 GiB. Object reference tasks are tightly concentrated around 485.10 steps because the remembered information is usually a stable referent, while temporal tasks include both short state checks and longer interaction chains.

\begin{table}[t]
\centering
\small
\renewcommand{\arraystretch}{1.08}
\setlength{\tabcolsep}{8pt}
\begin{tabular}{@{}lrr@{}}
\toprule
\textbf{Length bucket} & \textbf{Episodes} & \textbf{Share} \\
\midrule
0--199 & 50 & 3.33\% \\
200--399 & 199 & 13.27\% \\
400--599 & 687 & 45.80\% \\
600--799 & 108 & 7.20\% \\
800--999 & 290 & 19.33\% \\
1{,}000--1{,}199 & 110 & 7.33\% \\
1{,}200--1{,}499 & 31 & 2.07\% \\
1{,}500+ & 25 & 1.67\% \\
\bottomrule
\end{tabular}
\caption{
Distribution of expert demonstration lengths in action steps.
}
\label{tab:length_distribution}
\end{table}

Table~\ref{tab:length_distribution} shows that MEMOBench contains both compact and long horizon episodes. The largest bucket contains 687 demonstrations between 400 and 599 steps, while 456 episodes are at least 800 steps and 25 episodes are at least 1{,}500 steps. The shortest episodes are temporal top drawer checks, while the longest episodes come from the procedural buffered object swap task. This spread is useful for evaluation because the benchmark includes both localized memory decisions and extended sequences where the policy must preserve or revise information across many control steps.

\section{Data Collection and Quality Control}
\label{sec:appendix_data_collection}

MEMOBench tasks are instantiated from BDDL specifications on the LIBERO and robosuite stack. Each BDDL file defines regions, fixtures, objects, initial state predicates, a language prompt, and a final goal predicate. Each metadata file adds checkpoint language, operation labels, and executable checkpoint predicates. Expert demonstrations contain 927{,}444 control steps across 1{,}500 episodes, with a mean length of 618.30 steps and a maximum length of 2{,}012 steps. Evaluation rollouts start from randomized initial states within the BDDL region constraints, run 10 no operation stabilization steps, and then allow a 1{,}200 step policy horizon. The policy receives a third person RGB view, a wrist RGB view, proprioceptive state, and the task language prompt. The task definitions make the current observation insufficient for success, so a policy must preserve earlier references, state changes, or demonstrated routines across later decisions.

Each HDF5 task file stores one group per demonstration. For every episode, the required arrays are actions, states, third person RGB frames, and wrist RGB frames. The action tensor has shape $(T,7)$, the two RGB observation tensors have shape $(T,256,256,3)$, and the state tensor has shape $(T,D)$ with task dependent state dimension $D$. The released files contain state dimensions 38, 47, 51, 58, 71, 77, 84, 97, and 149. We verified that actions, states, and both image streams share the same episode length $T$ for every demonstration group.

Checkpoint labels are tied to simulator state rather than to model self reports. During a rollout, checkpoint predicates are evaluated after every policy step, and final task success is still judged by the task BDDL goal. The raw logger records whether each checkpoint predicate has ever been satisfied, while sequence aware summaries additionally apply an ordered prefix gate. This keeps checkpoint evaluation executable for reactive policies, context based policies, and explicit memory modules. It also makes process metrics a measurement of memory dependent checkpoint completion, not a direct probe of hidden memory states.

Quality control checks consistency between BDDL goals, checkpoint predicates, and operation labels. Because every task and checkpoint is executed through the same simulator predicate interface, missing or invalid objects, fixtures, regions, and predicates surface during environment instantiation or rollout evaluation rather than only in offline text. We also check the operation labels against the taxonomy in Section~\ref{sec:benchmark}: initial task information is labeled Storage, changed external state is labeled Update, and accumulated history is labeled Compression.

\section{Blind Annotator Agreement Study}
\label{sec:appendix_annotator_agreement}

To test whether annotators who did not author the released labels can apply the operation taxonomy consistently, we ran a blind relabeling study over all 84 checkpoint templates. Three annotators received identical written instructions together with the checkpoint language and the executable predicate of each template, and the released operation labels were hidden. Each annotator first completed a 10 item pilot that was excluded from scoring and then independently labeled the full set of 84 templates.

\begin{table}[t]
\centering
\small
\renewcommand{\arraystretch}{1.08}
\setlength{\tabcolsep}{4pt}
\begin{tabular}{@{}>{\raggedright\arraybackslash}p{0.56\linewidth}c@{}}
\toprule
\textbf{Statistic} & \textbf{Value} \\
\midrule
Fleiss' $\kappa$ (three raters) & 0.799 \\
Pairwise Cohen's $\kappa$ & 0.695 / 0.835 / 0.868 \\
Exact three rater agreement & 70/84 (83.3\%) \\
Majority vote vs.\ released labels & 71/84 (84.5\%) \\
Pooled vs.\ released labels & 203/252 (80.6\%) \\
Per rater vs.\ released labels & 77.4\% / 79.8\% / 84.5\% \\
\bottomrule
\end{tabular}
\caption{
Agreement statistics for the blind relabeling study over the 84 checkpoint templates. Pooled agreement counts each of the three annotator decisions per template, for 252 decisions in total.
}
\label{tab:annotator_agreement}
\end{table}

Table~\ref{tab:annotator_agreement} reports the agreement statistics. Agreement among the three raters is substantial, with Fleiss' $\kappa=0.799$ and pairwise Cohen's $\kappa$ between 0.695 and 0.868. All three raters assign the same label on 70 of 84 templates, and every disagreement is a one versus two split rather than a three way split. A majority vote over the three raters matches the released label on 84.5\% of templates.

\begin{table}[t]
\centering
\small
\renewcommand{\arraystretch}{1.08}
\setlength{\tabcolsep}{6pt}
\begin{tabular}{@{}lrrr@{}}
\toprule
\textbf{Released label} & \textbf{Ann. S} & \textbf{Ann. U} & \textbf{Ann. C} \\
\midrule
S (35 templates) & 100 & 5 & 0 \\
U (32 templates) & 2 & 90 & 4 \\
C (17 templates) & 0 & 38 & 13 \\
\bottomrule
\end{tabular}
\caption{
Pooled confusion matrix between released operation labels and annotator labels in the blind relabeling study. Each row pools the three annotator decisions per template, so rows sum to three times the template count.
}
\label{tab:annotator_confusion}
\end{table}

Table~\ref{tab:annotator_confusion} shows that the main mismatch is concentrated in the released Compression row. Annotators often assigned Update when the checkpoint text described a final physical state change rather than an explicit aggregate over multiple events. In total, 14 templates received any inter annotator disagreement, with 8 splits between Update and Compression and 6 splits between Storage and Update. The Update and Compression splits cluster at final two item drawer closures and at final procedural or counting checkpoints where a physical action also implies completion of a multi-step sequence. The Storage and Update splits occur when a compound task first switches to a new drawer, which some raters treated as a new memory record and others as an update within the same episode. These cases motivate an explicit adjudication rule. We label a checkpoint Compression only when its predicate or language requires an aggregate over multiple prior events or facts, Update when it is a later physical state change of an already tracked entity, and Storage when it establishes a new task relevant memory record for a distinct entity.

\section{Metric Formulation Details}
\label{sec:appendix_metrics}

This section expands the process metrics introduced in Section~\ref{sec:metrics}. For task $\tau$ evaluated over $N$ rollouts, let $\mathcal{C}_\tau=\{c_1,c_2,\ldots,c_K\}$ denote its ordered memory checkpoints. Each checkpoint has an operation label $o_k\in\{\texttt{S},\texttt{U},\texttt{C}\}$, which partitions checkpoints into $\mathcal{C}^{\texttt{S}}_\tau$, $\mathcal{C}^{\texttt{U}}_\tau$, and $\mathcal{C}^{\texttt{C}}_\tau$.

For rollout $d_i$, the verifier $v_i(c_k)\in\{0,1\}$ records whether checkpoint $c_k$ is completed under the simulator predicate rule from Appendix~\ref{sec:appendix_data_collection}. For sequence aware summaries, we also use the prefix indicator $\bar{v}_i(c_k)=\prod_{j=1}^{k}v_i(c_j)$, so an isolated or out of order satisfaction of $c_k$ cannot raise prefix completion. The paper tables report macro averages over applicable tasks and memory types. The rollout logger can additionally pool checkpoint records when summarizing diagnostic runs. For nonempty operation subsets, the task level metrics are
\begin{align}
    \mathrm{MSR}(\tau) &= \frac{1}{N} \sum_{i=1}^{N} \frac{1}{|\mathcal{C}^{\texttt{S}}_\tau|} \sum_{c \in \mathcal{C}^{\texttt{S}}_\tau} v_i(c), \label{eq:msr} \\
    \mathrm{MUR}(\tau) &= \frac{1}{N} \sum_{i=1}^{N} \frac{1}{|\mathcal{C}^{\texttt{U}}_\tau|} \sum_{c \in \mathcal{C}^{\texttt{U}}_\tau} v_i(c), \label{eq:mur} \\
    \mathrm{MCR}(\tau) &= \frac{1}{N} \sum_{i=1}^{N} \frac{1}{|\mathcal{C}^{\texttt{C}}_\tau|} \sum_{c \in \mathcal{C}^{\texttt{C}}_\tau} v_i(c). \label{eq:mcr}
\end{align}

Tasks without a given operation are marked as not applicable for that metric and are excluded from the corresponding macro average. MSR measures completion of predicates tied to newly observed task information. MUR measures completion of predicates tied to revised state after an external change. MCR measures completion of predicates tied to consolidated history. Together with SR, these metrics separate final task completion from checkpoint completion. Low SR with high process metrics points to action execution as a likely bottleneck, while a selective drop in MSR, MUR, or MCR localizes failure to the operation whose checkpoint predicates are never reached during the rollout.

\section{Detailed Memory Alignment Objectives}
\label{sec:appendix_alignment_objectives}

The alignment objectives in Section~\ref{sec:benchmark} add checkpoint text supervision to the RoboMME $\pi_{0.5}$ memory module~\cite{dai2026robommebenchmarkingunderstandingmemory}. Let $H \in \mathbb{R}^{B \times L \times d_m}$ denote encoded memory tokens for a batch and let $m \in \{0,1\}^{B \times L}$ denote the valid token mask. In the reported perceptual memory setting, $L=512$ and $d_m=2048$. The memory input uses one front camera view. The token budget is organized as 32 uniformly sampled history frames with 16 tokens per frame. Each encoded memory token is produced from SigLIP visual features and positional features.

The language target is the annotated \texttt{simple\_subgoal}. We encode every unique subgoal offline with a frozen SigLIP text encoder and cache text embeddings $T \in \mathbb{R}^{B \times d_t}$, where $d_t=1152$. The text encoder is used only to produce training targets. During rollout, the policy receives observations and memory tokens, but no alignment text target.

\paragraph{Semantic Memory Alignment.}
Semantic Memory Alignment gives the pooled memory state a direct language target. We compute a masked mean over memory tokens and project it into the text embedding space:
\[
\bar{h}_b =
\frac{\sum_{i=1}^{L} m_{b,i}H_{b,i}}
{\sum_{i=1}^{L} m_{b,i}},
\qquad
z_b = W\bar{h}_b + \beta.
\]
After L2 normalization, the objective minimizes the positive pair cosine distance:
\[
\mathcal{L}_{\mathrm{SMA}}
=
\frac{1}{B}\sum_{b=1}^{B}
\left(1-\hat{z}_b^{\top}\hat{T}_b\right).
\]
This objective is a lightweight semantic regularizer because it does not introduce batch negatives.

\paragraph{Contrastive Memory Alignment.}
Contrastive Memory Alignment makes the same pooled representation discriminative across subgoals. We compute memory to text logits
\[
Z_{i,j}=\frac{\hat{z}_i^\top \hat{T}_j}{\tau},
\qquad
\tau=\exp(\gamma),
\]
where $\gamma$ is a learnable log temperature initialized from $\tau=0.07$. Repeated subgoals within a batch are treated as positives rather than negatives. The positive target distribution is
\[
P_{i,j}=\mathbbm{1}\!\left[\hat{T}_i^\top\hat{T}_j > 0.99\right],
\qquad
Y_{i,j}=\frac{P_{i,j}}{\sum_k P_{i,k}}.
\]
The two retrieval directions are
\[
\mathcal{L}_{m\rightarrow t}
=
-\frac{1}{B}\sum_i\sum_j
Y_{i,j}\log \operatorname{softmax}(Z_{i,:})_j,
\]
\[
\mathcal{L}_{t\rightarrow m}
=
-\frac{1}{B}\sum_i\sum_j
Y_{j,i}\log \operatorname{softmax}(Z_{:,i})_j.
\]
The final contrastive loss is
\[
\mathcal{L}_{\mathrm{CMA}}
=
\frac{1}{2}\left(\mathcal{L}_{m\rightarrow t}
+\mathcal{L}_{t\rightarrow m}\right).
\]
This objective encourages each memory vector to retrieve its matching subgoal from batch candidates while avoiding false negative gradients for repeated labels.

\paragraph{Framewise Memory Alignment.}
Framewise Memory Alignment preserves the temporal layout of sampled history. We reshape memory tokens into $H' \in \mathbb{R}^{B \times F \times P \times d_m}$ and reshape the token mask into $A' \in \{0,1\}^{B \times F \times P}$, where $F=32$ sampled frame slots and $P=16$ tokens per frame. Each frame representation is a masked mean over the tokens in that frame:
\[
u_{b,f}=
\frac{\sum_{p=1}^{P} A'_{b,f,p}H'_{b,f,p}}
{\sum_{p=1}^{P} A'_{b,f,p}}.
\]
We then project each frame vector into the text embedding space:
\[
z_{b,f}=W_f u_{b,f}+\beta_f.
\]
The target text for a valid sampled frame is obtained by looking up \texttt{simple\_subgoal} at the same episode step. If the sampled frame index is not an annotated step, the lookup uses the next available annotated step. Frames beyond the last annotated step use the final annotation, and padded frame slots are masked. After projection and normalization, the frame loss is
\[
\ell_{b,f}=1-\hat{z}_{b,f}^{\top}\hat{T}_{b,f}.
\]
The valid mask $V_{b,f}$ combines frame validity and frame text validity. The reported framewise variant uses linear recency weighting:
\[
\alpha_{b,f}=0.5+0.5\frac{r_{b,f}}{\max(N_b-1,1)},
\]
where $r_{b,f}\in\{0,\ldots,N_b-1\}$ is the zero based rank of frame $f$ among the $N_b$ valid frames in episode $b$. The final objective is
\[
\mathcal{L}_{\mathrm{FMA}}
=
\frac{\sum_b\sum_f V_{b,f}\alpha_{b,f}\ell_{b,f}}
{\sum_b\sum_f V_{b,f}\alpha_{b,f}}.
\]

All three alignment losses are added only during training:
\[
\mathcal{L}_{\mathrm{total}}
=
\mathcal{L}_{\mathrm{action}}
+\lambda \mathcal{L}_{\mathrm{align}}.
\]
We use $\lambda=0.1$ for Semantic and Contrastive Memory Alignment and $\lambda=0.05$ for Framewise Memory Alignment. Evaluation remains unchanged and uses task success plus executable memory checkpoint predicates.

\section{Radar Summaries}
\label{sec:appendix_radar}

\noindent\begin{minipage}{\linewidth}
\centering
\includegraphics[width=0.485\linewidth]{content/image/table3_main_results_radar.pdf}\hfill
\includegraphics[width=0.485\linewidth]{content/image/table4_memory_alignment_radar.pdf}
\captionsetup{hypcap=false}

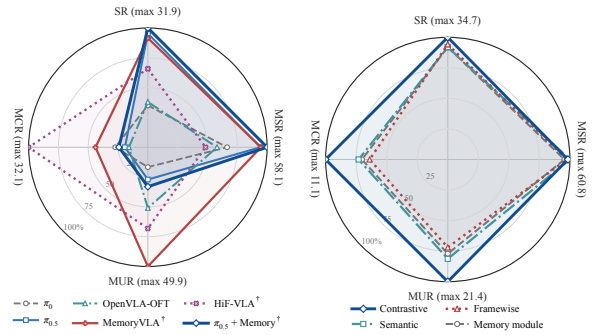
\captionof{figure}{Radar summaries for Table~\ref{tab:main_results} (left) and Table~\ref{tab:memory_alignment_results} (right). Each axis takes the best value over replan settings for each method and metric, then normalizes by the axis maximum.}
\label{fig:main_alignment_radar}
\end{minipage}

Figure~\ref{fig:main_alignment_radar} provides a compact view of the tradeoffs in Tables~\ref{tab:main_results} and~\ref{tab:memory_alignment_results}. This radar uses a best over replan rule, which differs from the mean over four replan settings used in the alignment discussion of Section~\ref{sec:experiments}. Under the radar rule, the $\pi_{0.5}$ memory module has the highest SR and MSR in the main comparison, MemoryVLA has the highest MUR, and HiF-VLA has the highest MCR. Under the same rule, Contrastive Memory Alignment has the highest SR, MUR, and MCR in the alignment comparison, while Semantic Memory Alignment has the highest MSR. These patterns support the main conclusion that current methods improve different memory operations rather than solving MEMOBench uniformly.

\section{MemER with Privileged Supervision}
\label{sec:appendix_memer}

\begin{table}[t]
\centering
\small
\renewcommand{\arraystretch}{1.14}
\setlength{\tabcolsep}{5pt}
\begin{tabular}{@{}l cccc@{}}
\toprule
\textbf{Memory type} & \textbf{SR} & \textbf{MSR} & \textbf{MUR} & \textbf{MCR} \\
\midrule
Temporal & 62.4 & 63.8 & 61.7 & 63.0 \\
Spatial & 50.4 & 84.3 & 52.8 & 77.6 \\
Object & 93.2 & 93.2 & -- & -- \\
Procedural & 19.6 & 20.4 & 21.0 & -- \\
\midrule
\rowcolor{gray!8}
\textbf{Average} & \textbf{56.4} & \textbf{65.4} & \textbf{45.2} & \textbf{70.3} \\
\bottomrule
\end{tabular}
\caption{
MemER results on MEMOBench. MemER receives checkpoint based keyframe and subtask supervision, so we report it as a privileged supervision reference row rather than a matched baseline for Table~\ref{tab:main_results}. Dashes mark memory types without the corresponding operation.
}
\label{tab:memer_results}
\end{table}

Table~\ref{tab:memer_results} reports MemER~\cite{sridhar2025memerscalingmemoryrobot} as a reference row with privileged supervision. We do not include MemER in the matched baseline comparison of Table~\ref{tab:main_results} because it is hierarchical. Its high level VLM selects keyframes and issues subtask instructions to a low level policy, which on MEMOBench consumes checkpoint based supervision from the same task suite used for evaluation. This makes the system stronger but not directly comparable to policies trained with matched supervision.

We run the official MemER recipe with checkpoint annotations used as keyframe and subtask supervision. The high level policy is Qwen2.5-VL-7B and the low level policy is $\pi_{0.5}$. Evaluation follows the same rollout protocol as the main results.

MemER improves average SR to 56.4, with average MSR 65.4, MUR 45.2, and MCR 70.3. The profile remains diagnostic. Spatial MUR stays at 52.8, and procedural SR and MUR stay at 19.6 and 21.0. Even with privileged checkpoint supervision, MEMOBench is far from saturated.

\section{Reproducibility Settings}
\label{sec:appendix_reproducibility}

Table~\ref{tab:reproducibility} reports the evaluation and training settings for the controlled $\pi_{0.5}$ memory alignment runs. These runs use the RoboMME memory module as the shared backbone~\cite{dai2026robommebenchmarkingunderstandingmemory}.

\noindent\begin{minipage}{\linewidth}
\centering
\small
\renewcommand{\arraystretch}{1.08}
\setlength{\tabcolsep}{3.5pt}
\begin{tabular}{@{}p{0.36\linewidth}p{0.56\linewidth}@{}}
\toprule
\textbf{Item} & \textbf{Setting} \\
\midrule
Train data & 1{,}500 expert demonstrations, 50 per task \\
Evaluation & 1{,}500 held out simulator rollouts, 50 per task \\
Evaluation seeds & base seed 7, episode seed $7+i$ \\
Rollout horizon & 10 stabilization steps and 1{,}200 policy steps \\
Image size & $256 \times 256$ render, $224 \times 224$ policy input \\
Action execution & 50 action predictions, replan steps $k \in \{5,20,35,50\}$ \\
Model size & $\pi_{0.5}$ has 3.2B parameters, the RoboMME memory module adds about 80M parameters in the modulator setting, and OpenVLA-OFT uses a Prismatic-7B backbone \\
Memory budget & 512 tokens from 32 frames and 16 tokens per frame \\
Memory view count & one front camera view \\
Memory feature composition & SigLIP visual features plus temporal and spatial positional features, encoded as 2048 dimensional tokens \\
Text target & \texttt{simple\_subgoal}, 1152 dimensional SigLIP text embeddings \\
Training random key & independently sampled for each run, not fixed to seed 7 \\
Checkpoint rule & final 80{,}000 update checkpoint for controlled alignment variants \\
Optimizer & AdamW, $\beta_1=0.9$, $\beta_2=0.95$, weight decay 0 \\
Learning rate & $5 \times 10^{-5}$, constant schedule, 5\% warmup \\
Batch and clipping & batch size 64, gradient clip norm 1.0 \\
EMA and saving & EMA decay 0.999, checkpoint every 10{,}000 updates \\
Alignment weights & 0.1 for Semantic and Contrastive, 0.05 for Framewise \\
Trainable parts & VLM expert, action expert, memory parameters, alignment heads \\
Frozen parts & SigLIP vision backbone with cached visual tokens \\
Compute & four A40 GPUs for about 3 to 4 days per $\pi_{0.5}$ memory run \\
\bottomrule
\end{tabular}
\captionsetup{hypcap=false}
\captionof{table}{Reproducibility settings for MEMOBench evaluation and the controlled $\pi_{0.5}$ memory alignment runs.}
\label{tab:reproducibility}
\end{minipage}

\end{document}